\documentclass{article}

\usepackage[preprint]{neurips_2024}

\usepackage[utf8]{inputenc} 
\usepackage[T1]{fontenc}    
\usepackage{hyperref}       
\usepackage{url}            
\usepackage{booktabs}       
\usepackage{amsfonts}       
\usepackage{nicefrac}       
\usepackage{microtype}      
\usepackage{xcolor}         
\usepackage{lipsum}         
\usepackage{CJKutf8}        
\usepackage{graphicx}       
\usepackage{amsmath}        
\usepackage{xspace}         
\usepackage{balance}        
\usepackage{makecell}
\usepackage{bm}
\usepackage{subcaption}
\usepackage{fancyhdr}
\usepackage{multirow}
\usepackage{bigdelim}
\usepackage{longtable}
\usepackage{booktabs}  
\usepackage{caption}   
\usepackage{wrapfig}

\title{YuLan-Mini: An Open Data-efficient  Language Model}

\author{%
Yiwen Hu\thanks{Team leaders.}~,~~ Huatong Song$^{*}$ \\
\textbf{Jia Deng,~~ Jiapeng Wang,~~ Jie Chen} \\
  \textbf{
  Kun Zhou,~~ Yutao Zhu,~~ Jinhao Jiang,~~ Zican Dong} \\
  \textbf{Wayne Xin Zhao}\thanks{Correspondence to Wayne Xin Zhao.},~~ \textbf{Ji-Rong Wen} \\
  Gaoling School of Artificial Intelligence \\
  Renmin University of China \\
  \texttt{batmanfly@gmail.com, jrwen@ruc.edu.cn}
}

\newcommand{\ie}{\emph{i.e.,}\xspace}

\newcommand{\eg}{\emph{e.g.,}\xspace}

\newcommand{\tabincell}[2]{\begin{tabular}{@{}#1@{}}#2\end{tabular}}

\newcommand{\ignore}[1]{}

\newcommand{\todo}[1]{\noindent\textcolor{darkblue}{\textbf{[{ToDo: }#1]}}}

\usepackage{tcolorbox}
\definecolor{darkorange}{RGB}{255, 140, 0}
\definecolor{darkblue}{RGB}{84, 112, 198}
\definecolor{lightgreen}{RGB}{145, 204, 117}
\definecolor{lightyellow}{RGB}{250, 200, 88}
\definecolor{lightred}{RGB}{238, 102, 102}
\definecolor{lightblue}{RGB}{115, 192, 222}

\definecolor{darkred}{rgb}{0.55, 0.0, 0.0}
\definecolor{navy}{rgb}{0.0, 0.0, 0.55}
\definecolor{darkgreen}{rgb}{0.0, 0.39, 0.0}

\newtcolorbox{promptbox}[2][Prompt]{
  colback=black!5!white,
  arc=5pt, 
  boxrule=0.5pt,
  fonttitle=\bfseries,
  title=#1, 
  before upper={\scriptsize}, fontupper=\fontfamily{ptm}\selectfont,
  colframe=#2, 
}

\newcommand{\newtt}[1]{%
    \ttfamily 
    #1
}

\begin{document}

\maketitle

\thispagestyle{fancy}
\fancyhead{}
\lhead{\raisebox{-0.88cm}{\includegraphics[height=0.6cm]{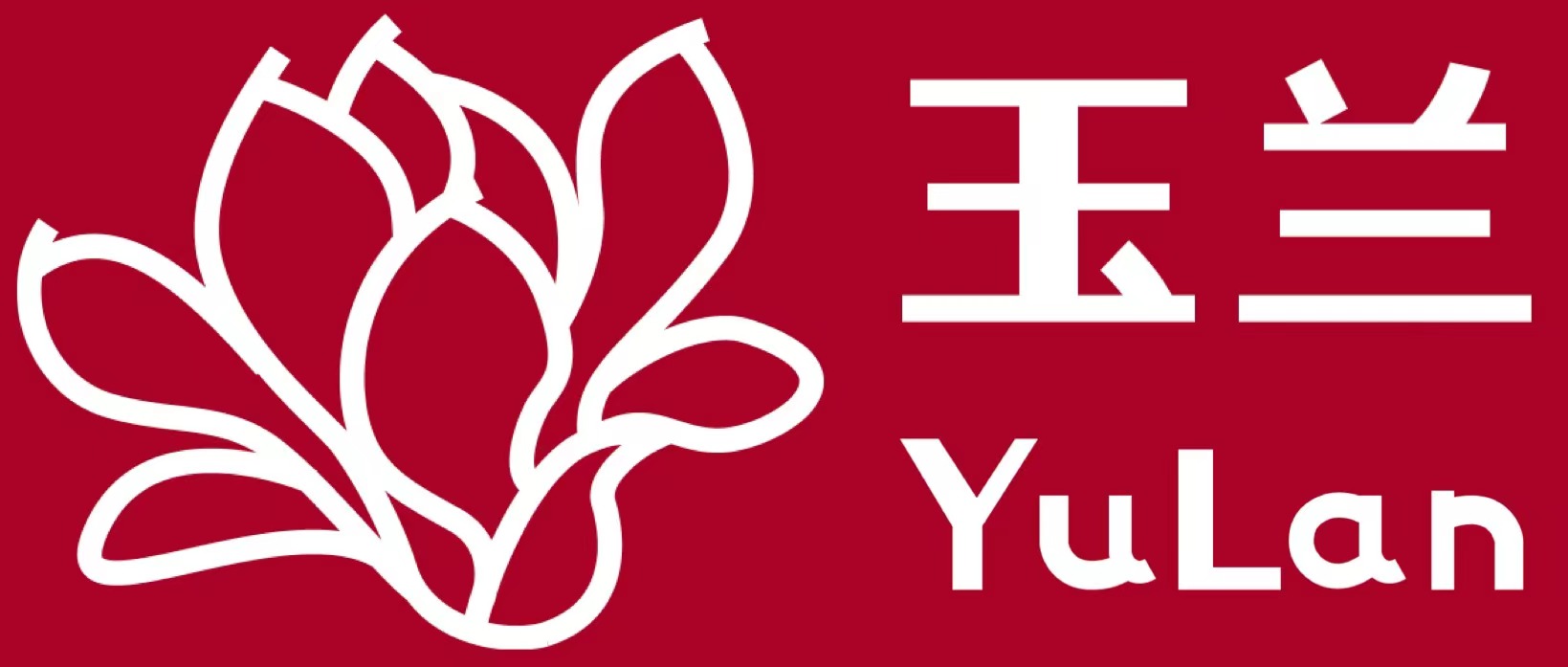}}\vspace{1mm}}

\begin{abstract}
Effective pre-training of large language models (LLMs) has been challenging due to the immense resource demands and the complexity of the technical processes involved. This paper presents a detailed technical report on YuLan-Mini, a highly capable base model with 2.42B  parameters that achieves top-tier performance among models of similar parameter scale. Our pre-training approach focuses on enhancing training efficacy through three key technical contributions: an elaborate data pipeline combines data cleaning with data schedule strategies, a robust optimization method to mitigate training instability, and an effective annealing approach that incorporates targeted data selection and long context training. Remarkably, YuLan-Mini, trained on 1.08T tokens, achieves performance comparable to industry-leading models that require significantly more data. To facilitate reproduction, we release the full details of the data composition for each training phase. 
Project details can be accessed at the following link: \url{https://github.com/RUC-GSAI/YuLan-Mini}.

\end{abstract}

\section{Introduction}\label{sec:introduction}

In recent years, large language models (LLMs)~\citep{zhao_survey_2023,llama-3-card,qwen2.5} have significantly advanced the frontier of AI technology. Unlike traditional machine learning methods, which are often specialized, LLMs excel across a diverse range of domains and tasks, showcasing their potential as versatile generalists.
Typically, LLMs are developed through a combination of pre-training and post-training techniques. 
It is widely recognized that pre-training is crucial for building the foundational capabilities of the LLMs~\citep{gpt-4-report,llama,deepseek-v2,qwen2.5-math}.

\begin{figure}[ht]
    \centering
    \includegraphics[width=0.95\linewidth]{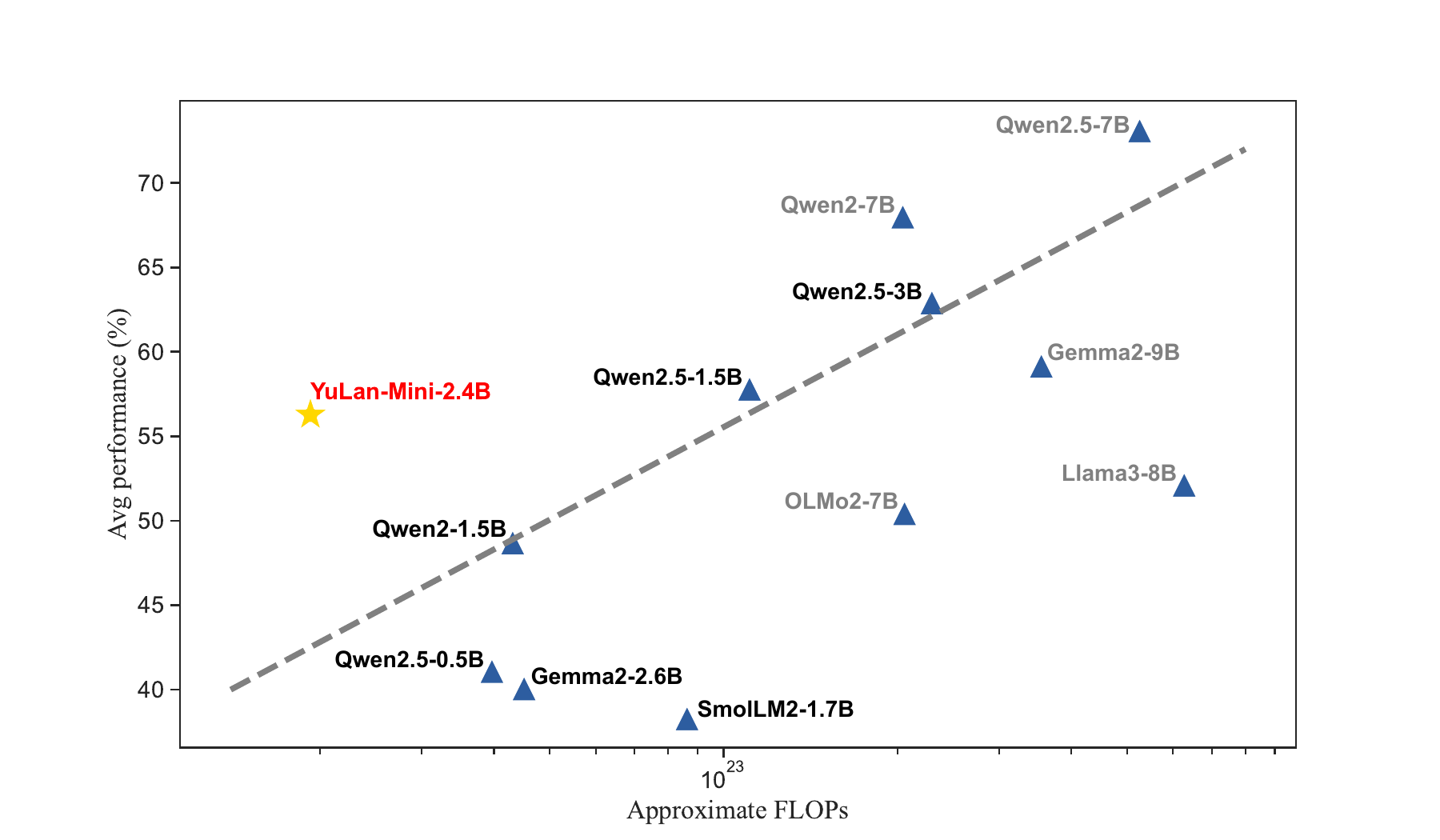}
    \caption{Performance comparison of YuLan-Mini against other \textbf{base models}, based on the average scores across eight benchmarks: GSM8K, MATH-500, HumanEval, MBPP, MMLU, ARC-Challenge, HellaSwag, and CEval.
   Floating Point Operations (FLOPs) are estimated using the scaling law formula $C=6ND$ proposed by~\cite{kaplan_scaling_2020}, where $N$ is the model size and $D$ is the size of the dataset.
   The models with a size  larger than 3B are plotted in gray.  
    }
    \label{fig:performance_compare}
\end{figure}

 

For transformer language models, the prevailing pre-training approach involves next-token prediction over large-scale unlabeled texts. Although this approach is conceptually straightforward, its implementation is technically complex. First, researchers must design an effective and efficient data pipeline to support pre-training, typically involving data collection, data cleaning, data mixing, and data curriculum. It is widely recognized that ``data'' is the most crucial element in enhancing model capabilities. Second, given that LLMs contain a vast number of parameters, the training process is challenging to stabilize and optimize. Common issues such as loss spikes or gradient explosions can occur during training, potentially leading to a failed process. 
Despite the availability of extensive model checkpoints released by industry companies, the core technical details often remain undisclosed in public reports. As a result, we know very little about how top-tier language models are developed in the industry.

Fortunately, the research community has made significant efforts to enhance the availability  of data resources and the  openness of training methodologies for LLM pre-training~\citep{allal2024SmolLM2,groeneveld_olmo_2024,map-neo}. First, well-curated datasets have been released to support the data preparation required for LLM pre-training. Additionally, various open research publications have documented the overall training procedures, providing a foundational understanding of LLM pre-training~\citep{hu_minicpm_2024,dclm-report}. These contributions offer basic technical approaches and essential resources for pre-training an LLM.





Despite these advancements, \emph{open LLMs}—those with fully disclosed technical details—still face two main limitations. First, most of these models tend to underperform compared to their industry counterparts due to constraints in data and computational resources. While some open LLMs achieve performance levels comparable to industry models, they also require similar amounts of resources, making them difficult to replicate within the research community.
Therefore, developing competitive LLMs with limited training resources remains a challenge, particularly in university-level laboratories.
Given these constraints, we aim to advance the openness of LLM pre-training by significantly enhancing both the performance ceiling and training efficiency of open models. Specifically, we focus on developing relatively small-scale language models, that is, models with a parameter scale ranging from 1B to 3B with a restricted compute budget.\footnote{Some papers refer to pre-trained language models of this size as ``small language models''. However, due to the technical similarities, we continue to refer to them as large language models in this paper.}  Our goal is to build a small yet powerful language model using only publicly available data, while sharing experiences or insights on improving training efficiency with limited computational resources.
In this paper, we present a comprehensive technical report on the development of a highly capable 2.42B-parameter language model (the base model) that achieves top-tier performance among models of similar parameter scale. 
Building upon the transformer architecture, we devise a data-efficient pre-training approach for LLMs. Our approach includes three major contributions to enhance training efficacy: 
(1) an elaborately designed \emph{data pipeline} that combines data cleaning with data schedule strategies; 
(2) a systematic \emph{optimization method} that can effectively mitigate training instability; 
(3) an effective \emph{annealing approach} that integrate targeted data selection and long context training. 
We explore a variety of techniques to enhance the performance of YuLan-Mini. In particular, we extensively leverage synthetic data for model training, including o1-like long-thought data. Additionally, we investigate multiple factors that may contribute to training instability. We provide two versions of the checkpoints, supporting 4K and 28K contexts, respectively.\footnote{Due to resource constraints, we were only able to train a model with up to 28K context.}


To demonstrate the effectiveness of our pre-trained base model, we conduct extensive experiments on a variety of benchmarks, and compare it with a few competitive base models from both research and industry.
Experimental results show that our base model, YuLan-Mini, can achieve very promising results among these compared models. 
For instance, it (the 28K version) achieves scores of 37.80 on MATH-500 (four-shot), 64.00 on HumanEval (zero-shot), and 49.10 on MMLU (five-shot). Figure~\ref{fig:performance_compare} presents a comparison of YuLan-Mini with other industry models.\footnote{We select eight popular benchmarks to cover math, coding, general, and language capabilities. This figure is primarily intended to illustrate the training efficacy of our model, rather than to serve as an accurate  capacity ranking of existing models.}



\textbf{To facilitate reproduction, we report the complete training details for YuLan-Mini, and also release the data composition for all training phases (Appendix~\ref{data-all-data}).} 
More supporting resources can be accessed at our project link: \url{https://github.com/RUC-GSAI/YuLan-Mini}. 

\section{Overall Pre-Training Configuration}\label{sec:pretraining-configuration}

In this section, we will provide an overview of the pre-training configuration, introducing its key components and the algorithms involved in the process. For a more detailed discussion of the major contributions made in this work, please refer to Section~\ref{sec:training_stability}, Section~\ref{sec:data_pipeline}, and Section~\ref{sec:decay}.



\subsection{Model Architecture}
Our model is based on a decoder-only transformer with a tall and narrow architecture, inspired by previous studies~\citep{liu_mobilellm_2024,hu_minicpm_2024}. It comprises a total of 2.42B parameters, of which 2.23B are non-embedding parameters.
The hyperparameter configurations for our model architecture are provided in Table~\ref{tab:architecture}. Additionally, we re-parameterize {each weight matrix of different modules} with an extra learnable parameter~\citep{nishida_initialization_2024}, enhancing the model's training stability (discussed in Section~\ref{sec:training_stability}). Next, we briefly introduce the main components in our architecture.


\paragraph{Embedding tying} We utilize embedding tying~\citep{press-wolf-2017-using} to reduce the model's parameter size and stabilize training. In our preliminary experiments, we find that sharing the embedding and unembedding matrices improves model convergence. Furthermore, when these matrices are not shared, they often necessitate different initialization strategies, which we will discuss in Section~\ref{sec:training_stability}.


\paragraph{Pre-RMSNorm}
Layer normalization~(LN) has been shown to enhance numerical stability and accelerate learning speed~\citep{ba2016layernormalization}.
We integrate Pre-LN into our model architecture to improve convergence stability and speed compared to Post-LN~\citep{xiong_layer_2020}. Regarding the form of normalization, we opt for \texttt{RMSNorm} over the conventional \texttt{LayerNorm}, as it conserves CUDA memory while attaining a comparable effect~\citep{zhang_root_2019}.


\paragraph{SwiGLU} Our model introduces non-linearity using a gated linear unit (GLU) with the Swish activation function, known as SwiGLU~\cite{shazeer_glu_2020}. This method effectively captures complex data relationships and has proven to be effective in relatively small language models, as demonstrated by \citep{liu_mobilellm_2024}.



\paragraph{Attention mechanism} We adopt the grouped-query attention (GQA,~\cite{ainslie_gqa_2023}), which enables the model to reduce KV cache usage while maintaining high performance. Specifically, we employ 30 heads for query attention and 6 groups for key-value heads. We opt not to make the KV head size divisible by 8 since small language models rarely require tensor parallelism during inference.


\paragraph{Rotary Embedding}\label{sec:rope} We adopt rotary positional embedding~(ROPE) to capture the positional information in our model, since it integrates absolute and relative positioning in an unified way. 
During the stable training stage, we set the parameter $\theta$ to 10,000, and increase it to $49\,000$  during the annealing stage to extend the context length to 28,672 (28K) tokens using adjusted base frequency (ABF).



\begin{table}[t]
    \small
    \caption{Hyperparameter settings of diffrent models. $r_\text{ffn}$ is the ratio of the feed-forward network's hidden size to the model's hidden size. The definition of the symbols is available at Table~\ref{tab:stability_recipe_explain}}
    \begin{center}
    \begin{tabular}{lccccc}
    \toprule
    \textbf{Model} & {$n_\text{layers}$} & {$d_\text{model}$} & {$r_\text{ffn}$} & {$n_\text{heads}$} & {$n_\text{kv\_heads}$} \\
    \midrule
    LLaMA-3.2-3B &28 &3,072 & 2.7 &24&8  \\
    Phi-3-mini-4k-instruct&32 &3,072 &2.7 &32&32 \\
    MiniCPM-2B &40&2,304 &2.5 &36&36  \\
    MiniCPM3-4B &62  & 2,560&2.5 &40&40  \\
    Qwen2.5-1.5B &28 &1,536  & 5.8&12 &2  \\
    MobileLLM-1B &54&1,280  &2.8 &20 &5  \\
    \midrule
    YuLan-Mini & 56 & 1,920 & 2.5 & 30 & 6 \\
    \bottomrule
    \end{tabular}
    \end{center}
    \label{tab:architecture}
\end{table}


\subsection{Tokenizer}
Tokenization is a critical preprocessing step that splits input text into sequences of tokens. Below, we provide details of our tokenizer.

\paragraph{Vocabulary size} 
Generally, the vocabulary size should be chosen to balance its effects on the model's parameter size and efficiency. 
We adopt the three approaches proposed by~\cite{dagan_getting_2024} to balance the compute budget and vocabulary capacity, yielding a final vocabulary size of around 99,000.
For simplicity, we reuse the Byte Pair Encoding (BPE) tokenizer of MiniCPM~\citep{hu_minicpm_2024}. Specifically, we truncate the vocabulary by applying the corresponding BPE merge rules to reduce the number of tokens.  
We also heuristically remove rare domain-specific tokens, while add some reserved tokens in the vocabulary. 
The statistics of the modified vocabulary and the compression rate are shown at Table~\ref{tab:tokenizer}. 
The test set for the tokenization experiments is sourced from a diverse array of datasets, as detailed in Section~\ref{sec:monitor}.
Overall, our tokenization method achieves a well-balanced compression rate across different domains.

\begin{table}[t]
    \centering
    \small
    \caption{Compression rate of different tokenizers. Higher values indicate more effective compression.}
    \begin{center}
    \begin{tabular}{lrcccc}
    \toprule
    \textbf{Tokenizer} & \textbf{Vocabulary Size} & \textbf{Web} & \textbf{Chinese} & \textbf{Math} & \textbf{Code} \\
    \midrule
    Gemma2-2B & 256,000 &4.928 &  3.808 &2.865 & 3.309\\
    Qwen2.5 & 151,936 & \underline{4.935} & 3.956 & 2.890 & \underline{3.881} \\
    LLaMA-3.1 & 128,000 & \textbf{4.994} & 3.263& \textbf{3.326} & \textbf{3.911} \\
    MiniCPM-2.4B & 122,753 &4.753 & \textbf{4.273} & \underline{2.739} &3.052 \\
    Phi-3.5-mini & 100,352 & 4.311 & 1.914 & 2.654 & 3.110 \\
    MiniCPM-1.2B & 73,440 &  4.631 & 4.042 & 2.696 & 3.017 \\
    \midrule
    YuLan-Mini & 99,000 & 4.687 & \underline{4.147} & 2.716 & 3.033 \\
    \quad + Dropout & 99,000 &4.687 &4.146 &2.715 &3.031 \\
    \bottomrule
    \end{tabular}
    \end{center}
    \label{tab:tokenizer}
\end{table}


\paragraph{BPE-dropout} Existing sub-word tokenization methods prevent the language models from understanding the alphabetic composition of a token.
To mitigate this issue, BPE-dropout~\citep{provilkov_bpe-dropout_2020} has been proposed to help the model better learn the internal representation of a token, enabling it to more effectively capture possible subwords within a word.   
 Specifically, we use a relatively low dropout rate of 0.2, and applying the dropout method results in only a slight increase in the number of tokens (0.07\%), as shown in Table~\ref{tab:tokenizer}. 
 


\paragraph{Digit tokenization}  Digit tokenization plays a crucial role in mathematical tasks, including numerical calculation and complex reasoning. We follow the common practice of splitting numbers into individual digits~\citep{deepseek-ai_deepseek_2024,yang_baichuan_nodate}. Although other methods, such as three-digit tokenization, may achieve higher compression rates, using individual-digit tokenization typically leads to improved numerical calculation accuracy~\citep{wang_tokenization_2024}.



\subsection{Training Data Preparation}
Data serves as the foundation for developing the model's capabilities, and we employ specially designed strategies for collecting and preparing the training dataset. Next, we briefly describe the general procedure for data preparation. A more detailed and comprehensive description of the data pipeline is provided in Section~\ref{sec:data_pipeline}.

\paragraph{Data collection and selection}


To ensure reproducibility, our pre-training data is primarily sourced from open-source pretraining datasets and synthetically generated data. {The main open-source datasets include {FineWeb-Edu}~\citep{lozhkov2024fineweb-edu}, {the-stack-v2}~\citep{lozhkov_starcoder_2024}, {open-web-math}~\citep{openwebmath}, {Chinese-FineWeb-Edu}~\citep{OpencsgChinesefinewebeduDatasets}, and {OpenCoder-LLM}~\citep{huang_opencoder_2024}.}  The entire pre-training dataset has undergone rigorous preprocessing, with 1.08T tokens for training. Among them are 481B English web data, 138B general English knowledge, 227B code pre-training data, 16.7B code instruction data, 93.8B mathematics pre-training data, 15.5B mathematics instruction data, and 108B Chinese data. 




\paragraph{Data schedule}  
Using the WSD scheduling method~\citep{hu_minicpm_2024}, the training process is divided into three main stages: warmup, stable training, and annealing. The warmup stage uses 10B tokens, the stable training stage utilizes 990B tokens, and the annealing stage uses 80B tokens. To better manage the training process, we divide the entire training trajectory into 27 consecutive \emph{curriculum phases}, each consisting of 40B tokens.  When transitioning between these curriculum phases, the dataset proportions are slightly adjusted based on the model's performance on various benchmarks and the perplexity (PPL) of validation texts. However, the internal data distribution of each curriculum phase cannot be modified once it has been scheduled for training.  During the annealing stage, the proportion of instruction data and long context data is increased. 

\subsection{Model Optimization}

For model optimization, hyperparameters are crucial for training stability and model performance. 

Specifically, we adopt the WSD learning rate scheduler~\citep{hu_minicpm_2024}.
Maintaining a constant learning rate during the stable training stage eliminates the necessity to specify an ending step, as required by the cosine scheduler. This approach facilitates continuing pre-training from the last checkpoint during stable training. It also allows for more flexible data preparation: we can prepare the data while the preceding curriculum phase is running. 
Additionally, we estimate an optimal annealing ratio of 8\% for the stable training stage using the scaling law of learning rate annealing~\citep{tissue_scaling_2024}. 

For training stability, we combine a parameter initialization approach akin to $\mu$P~\citep{dey_cerebras-gpt_2023,hu_minicpm_2024,yang_tensor_2022} with WeSaR re-parameterization~\citep{nishida_initialization_2024}, using a relatively large global learning rate of 0.01.
The rationale behind adopting a large learning rate is the expectation that the model will possess greater potential for enhancement during the annealing stage. 
We set the AdamW hyper-parameters 
as follows: $\beta_1=0.9, \beta_2=0.95, \epsilon=10^{-15}$,  with the \texttt{weight\_decay} of $0.1$ and the \texttt{z-loss} coefficient of $10^{-4}$~\citep{brebisson_z-loss_2016}. We use a variance of $5\times10^{-5}$ for initialization. As found by~\cite{wortsman_small-scale_2023}, extending the warm-up ratio enhances training stability, so we linearly warm up the model over 10B tokens.  We use a batch size of 4.12M tokens with a sequence length of 4,096, extending the context length during the annealing stage while keeping the total token count in the batch size unchanged. We avoid using gradient accumulation to prevent numerical precision error of \texttt{bfloat16}.  Detailed analysis of training stability can be found in Section~\ref{sec:training_stability}.





\subsection{Training Infrastructure}

We build a simple yet efficient training framework based on the HuggingFace  \texttt{Trainer} and other open-source libraries (\texttt{DeepSpeed}, \texttt{flash-attention}, and \texttt{liger-kernel}). 

Specifically, we first use ZeRO-1~\citep{rajbhandari_zero_2020} data parallelism provided by \texttt{DeepSpeed} intergration and then switch to ZeRO-2 after confirming that it does not cause training divergence in our model.\footnote{\href{https://github.com/microsoft/DeepSpeed/issues/6351}{https://github.com/microsoft/DeepSpeed/issues/6351}}
We also leverage Flash Attention~\citep{dao_flashattention_2022, dao_flashattention-2_2023} and a triton kernel library \texttt{liger-kernel}~\citep{hsu_liger_2024} to accelerate training processes. By employing fused kernels, we achieve a 30\% reduction in training time and up to 70\% savings in CUDA memory.\footnote{Fused kernels include: SelfAttention, RMSNorm, RoPE, SwiGLU, FusedLinearCrossEntropy, and AdamW. \texttt{torch.compile} is also enabled in our implementation.} 
We further optimize the balance between CUDA memory usage and training time by adjusting the number of layers through the activation checkpointing function. 
For enhanced training efficiency, we use \texttt{bfloat16} precision for both model parameters and NCCL communications. 
The model's FLOPs utilization (MFU) is estimated at $51.57\%$.


Regarding the hardware setup, we initially employ a 56 A800-GPU cluster managed by the SLURM system~\citep{yoo_slurm_2003}. We later reduce the number of GPUs to 48 by transitioning the distributed optimizer to a universal checkpoint~\citep{lian_universal_2024}. To maximize device utilization, we perform tokenization and packing asynchronously. Given the modest size of our cluster, the likelihood of encountering NCCL failures is relatively low. Therefore, after assessing the advantages and disadvantages, we decide to store a checkpoint every hour and implement automatic restarts.


For efficient evaluation, we utilize LLMBox~\citep{tang_llmbox_2024} to integrate vLLM~\citep{kwon_efficient_2023} for generative tasks and employ KV cache scheduling for multiple-choice tasks. For a detailed description of the evaluation setup and results, please refer to Section~\ref{sec:decay}.


\section{Training Stability}\label{sec:training_stability}


\begin{figure}[t]
    \centering
    \begin{subfigure}[b]{0.49\textwidth}
        \centering
        \includegraphics[width=\textwidth]{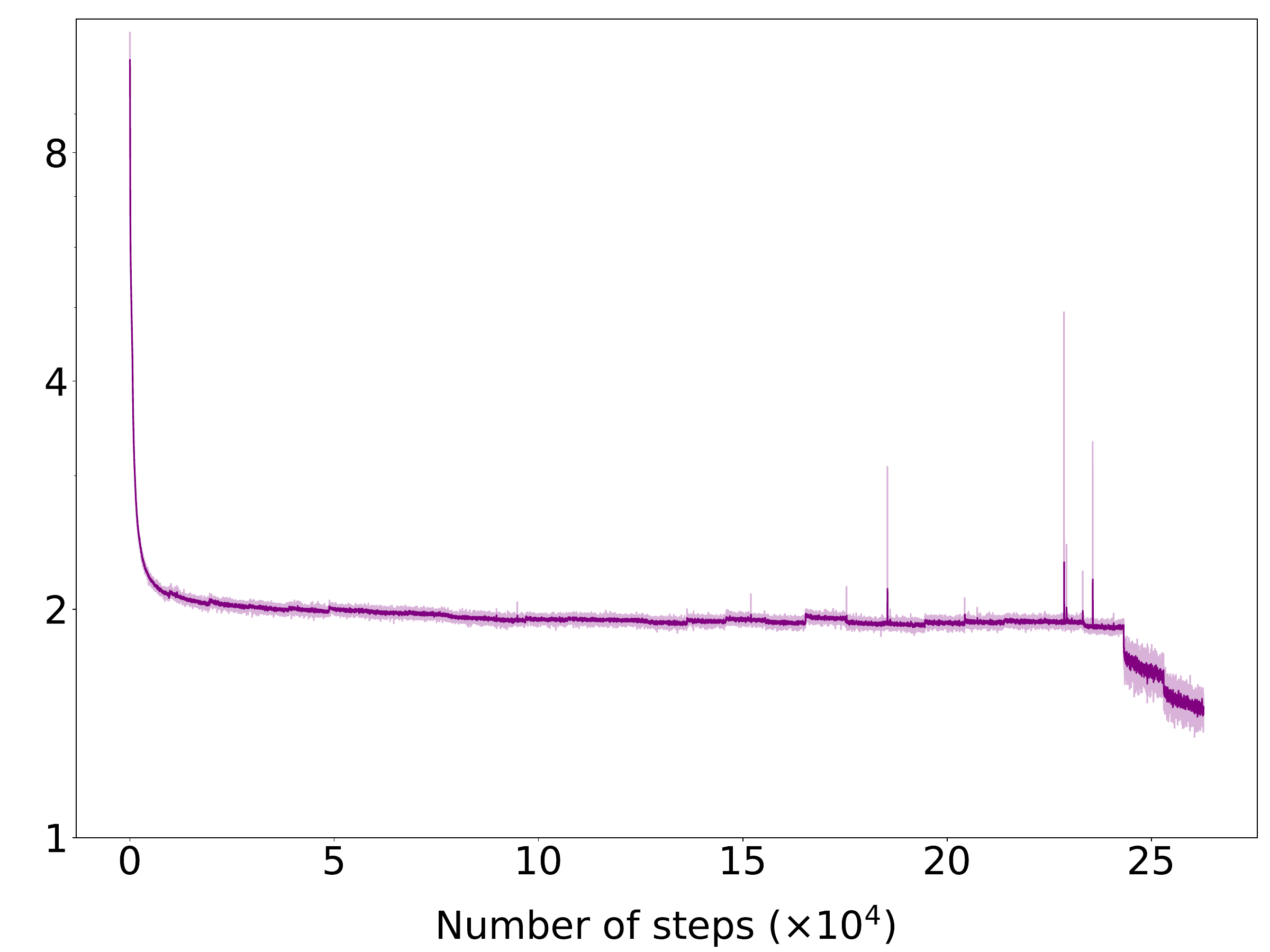}
        \caption{Training loss.}
        \label{fig:syn_compare_subfig1}
    \end{subfigure}
    \begin{subfigure}[b]{0.49\textwidth}
        \centering
        \includegraphics[width=\textwidth]{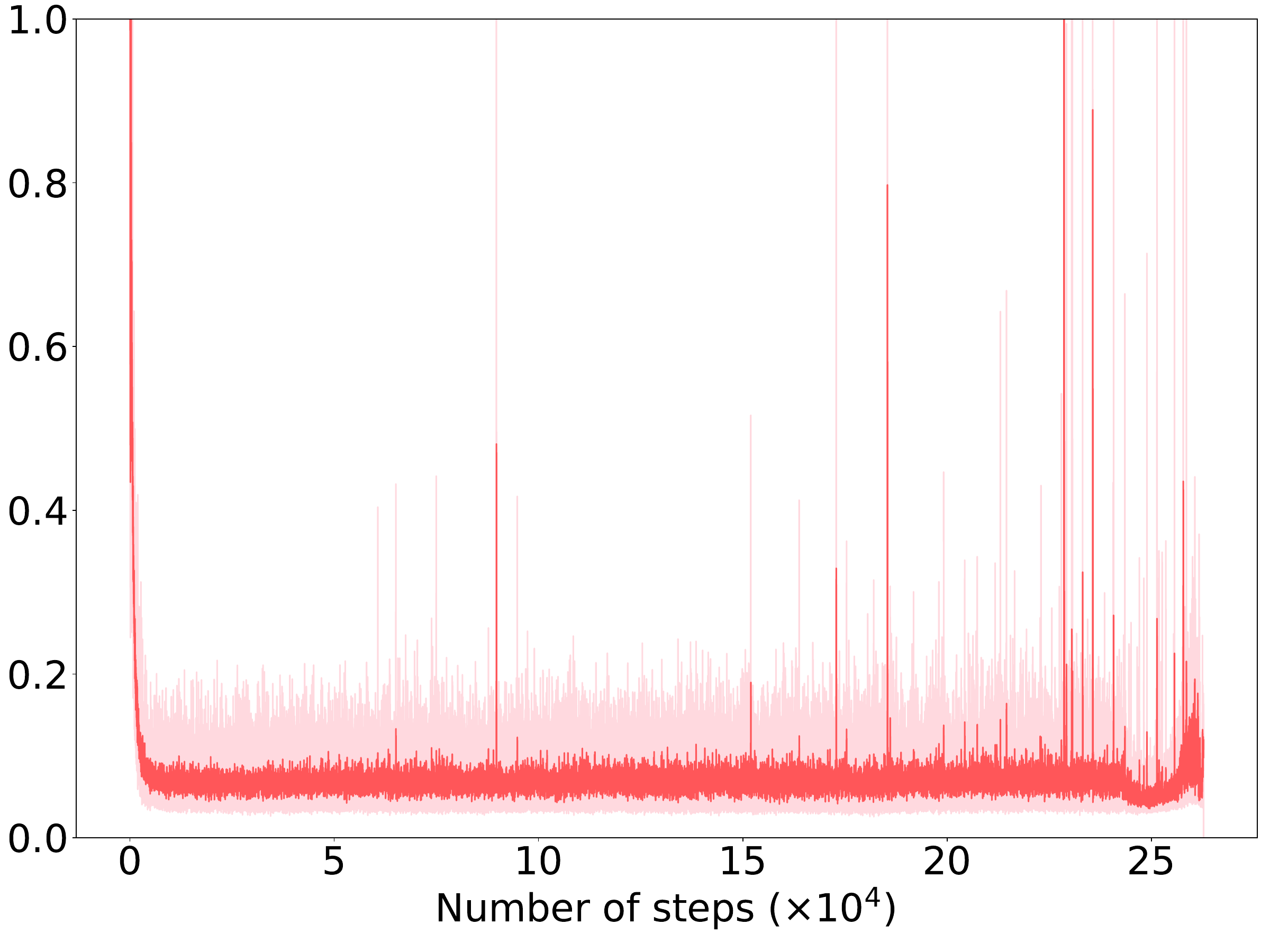}
        \caption{Gradient norm.}
        \label{fig:syn_compare_subfig2}
    \end{subfigure}
    \caption{Training loss and gradients during pre-training process.}
\label{fig:syn_compare}
\end{figure}

Training stability is crucial for the efficient pre-training of LLMs. Under normal conditions, loss trajectories are expected to decrease smoothly and remain near their anticipated values consistently, such as those predicted by the scaling law curve, even in the presence of minor perturbations. However, models that are improperly initialized or trained with unsuitable architectures or hyper-parameters may experience marginal stability or outright instability. In a marginally stable network, even a minor abnormal perturbation from the data can push it into a non-steady state. If the network can self-correct, it will experience temporary loss spikes; otherwise, training may diverge.

Training stability issue exists even in training relatively language models. For example, as observed empirically by~\cite{wortsman_small-scale_2023}, under the same learning rate, the smaller the model, the larger the order of magnitude of the model logits, which is a typical factor for training instability. 
General methods, such as replacing the data that leads to spikes or reducing the learning rate, usually only palliate superficial problems.

Despite significant efforts in the literature to mitigate training instability~\citep{takase_spike_2024,yang_tensor_2022,wortsman_small-scale_2023}, prior studies typically focus on individual techniques or conduct relatively small-scale experiments. There remains a lack of systematic investigation into the effects of various potential techniques in large-scale pre-training experiments. During our pre-training process, we encounter severe training instability issues, prompting us to conduct an in-depth study on how to address this problem effectively. Our primary approach involves combining a $\mu$P-like initialization~\citep{CerebrasGPT} with a re-parametrization method~\citep{nishida_initialization_2024} to adjust the learning rate and stabilize training. In the following, we present a detailed approach for maintaining training stability.

\subsection{Exploring the Hidden States Variability and Training Instability}

To effectively mitigate training instability, it is crucial to examine potential indicators of abnormal states.
Generally, loss, gradient, and hidden states are three interconnected factors that reflect the dynamics of training.
Among these indicators, loss provides surface-level clues about instability, while gradients and hidden states often reveal deeper underlying factors that contribute to a pathological state.
We begin by conducting a preliminary experiment to assess the impact of different indicators on training instability. We then analyze the potential reasons theoretically.


\subsubsection{Preliminary Experiment on Indicators}
We conduct a preliminary experiment to showcase the effects of tracking our indicators based on  hidden states during training.

\paragraph{Training setup}

Since it is resource-intense to perform extensive experiments on our model, we  explore the training dynamics by conducting surrogate experiment with a small proxy model of 0.2B with similar architecture. We employ a relatively large learning rate of 0.01, to expose potential instabilities within the model.
{We keep this baseline model setup in the subsequent experiment, which we elaborate on in Appendix~\ref{sec:appendix-training-stability}.} 
Specifically, our optimization goal is to achieve optimal performance while ensuring that the training process does not result in divergent loss or an increasing trend in gradient norm.

\paragraph{Indicators setup}
In large-scale training, distributed optimizers are often used, which means that the gradients of different modules may be distributed across various data parallel ranks. This distribution makes it inefficient to directly obtain the gradients. As a result, we primarily track each module's weight matrix and hidden states (\ie their outputs). Specifically, we record the mean and variance of the weights and hidden states, as well as the root mean square (RMS), which is calculated using the follow formula
$\text{RMS}=\sqrt{\text{Var} + \text{Mean}^2}$. 
Note we consider the outputs of various modules in the transformer (\ie \texttt{FFN}, \texttt{Attention}, \texttt{RMSNorm}) as hidden states. 

\begin{figure}[t]
    \centering
    \begin{subfigure}[b]{0.31\textwidth}
        \centering
        \includegraphics[width=0.99\textwidth]{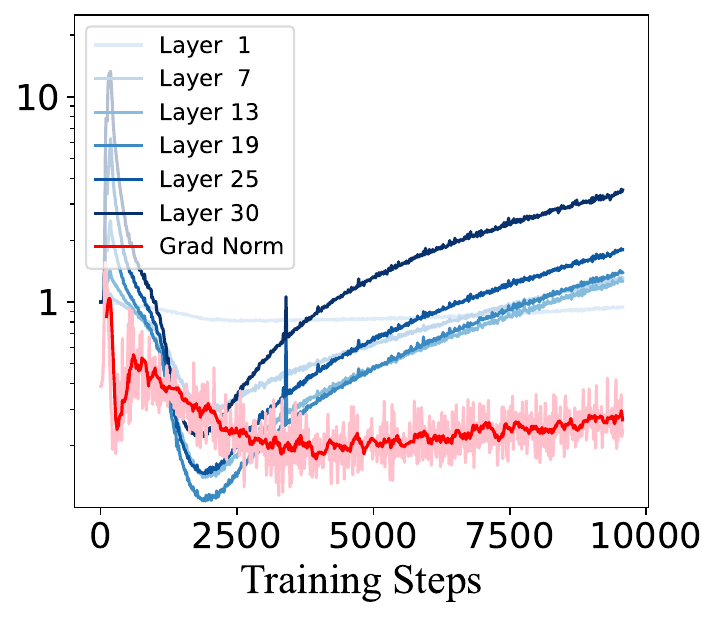}
        \caption{Exploding hidden states.}
        \label{fig:exploding_var}
    \end{subfigure}
    \hfill
    \begin{subfigure}[b]{0.315\textwidth}
        \centering
        \includegraphics[width=0.99\textwidth]{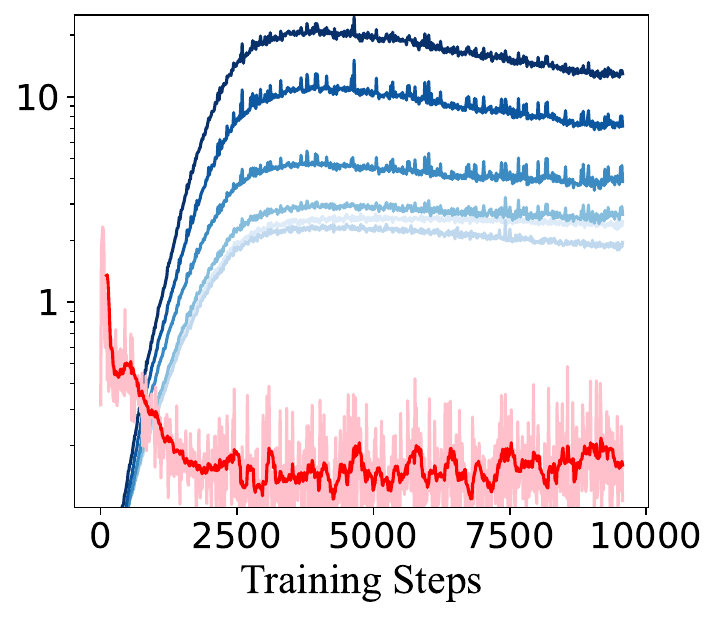}
        \caption{Convergent hidden states.}
        \label{fig:convergent_var}
    \end{subfigure}
    \hfill
    \begin{subfigure}[b]{0.315\textwidth}
        \centering
        \includegraphics[width=0.99\textwidth]{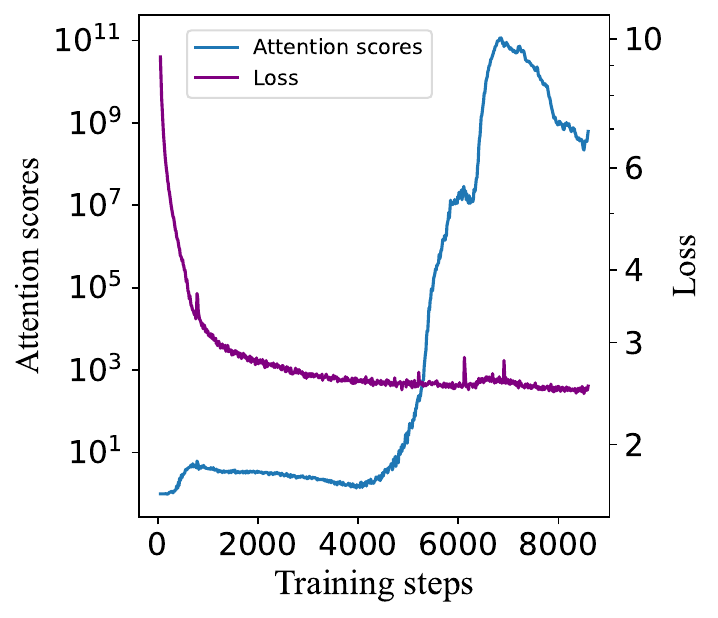}
        \caption{Loss prediction failure.}
        \label{fig:loss_prediction_failure}
    \end{subfigure}
    \caption{Comparison of training dynamics between divergent and convergent trial. The $y$-axis denotes the value of the hidden states variance and gradient norm on a log-scale. Both trials have consistent loss, but different trends of hidden states variance and gradient norm.}
\label{fig:syn_compare}
\end{figure}

\paragraph{Empirical findings}
Figure~\ref{fig:exploding_var} illustrates our findings on the relationship between training instability and exploding hidden states. Across all layers, there is a consistent upward trend in both hidden states and gradient norms as the number of training steps increases. Interestingly, these intermediate trends are difficult to detect in the early stages of pre-training when focusing solely on the loss. 
Moreover, in addition to the temporal dimension, indicators related to hidden states also show an increasingly divergent trend across the depth of the model (\ie as the number of layers increases). The ratio of the variance of the last layer to that of the first layer grows linearly, indicating a potential future explosion in hidden states and gradient norms. 
From these results, we can see that hidden states play a significant role in affecting training stability. 
Based on these observations, our core idea is to monitor and adjust hidden states to maintain training stability.

In addition to monitoring hidden states, other indicators can help detect abnormal optimization issues. In our experiments, we also examine another stability indicator known as token embedding variability (TEV)~\citep{chung_stable_2024}, which measures the variance of data entries within a vector.   
By comparing the TEV of two input vectors, we can quantify how differently they are distributed. 
If the input vectors follow distributions that are significantly different from each other, the output vectors may of course exhibit large fluctuations. 
These output vectors are essentially a special kind of hidden states and therefore can detect training instability. However, since we are already recording hidden states in our experiments, we did not use TEV as an additional indicator.

\subsubsection{Theoretical Analysis and Empirical Evidence on Exploding Hidden States
}\label{sec:hidden_states_theoretical_analysis}
In this part, we first formalize the hidden states of transformer, and then derive three potential reasons for exploding hidden states theoretically. We also showcase corresponding empirical evidence, to verify the effectiveness of using hidden states as a training stability indicator. 

\paragraph{Formalization of hidden states} To better analyze  how hidden states can be utilized to indicate training stability, we begin by formally defining the hidden states in our model. Since Yulan-Mini is a multi-layer transformer model, and the hidden states of the $l$-th layer, denoted by $\bm{z}^l\in\mathbb{R}^d$, can be specified  as follows: 
\begin{align*}
    \bm{z}^l &= \bm{y}^l + \texttt{FFN}(\texttt{RMSNorm}(\bm{y}^l)), \\
    \bm{y}^l &= \bm{x}^l + \texttt{MHA}(\texttt{RMSNorm}(\bm{x}^l)),
\end{align*}
where $\bm{x}^l\in\mathbb{R}^d$ denotes the input of the $l$-th layer. 
With sub-layer inputs $\bm{u}=\texttt{RMSNorm}(\bm{x}^l)$ and $\bm{v}=\texttt{RMSNorm}(\bm{y}^l)$, the \texttt{FFN} and \texttt{MHA} are defined as: 
\begin{align}
    \texttt{FFN}(\bm{u})&=\mathbf{W}_2\mathcal{F}(\mathbf{W}_1\bm{u}), \label{eq:ffn}\\
    \texttt{MHA}(\bm{v})&=\text{concat}_{i=1}^{h}[\texttt{head}_i(\bm{v})]\mathbf{W}_o. \label{eq:mha}
\end{align}
{It is worth noting that we focus on discussing a simplified version of the FFN layer above. Modern LLAMA-like architectures typically utilize GLU-style non-linearities, which we discuss in Section~\ref{sec:stability_initialization}. 
} 
Building on this definition, we specifically focus on the change in the variance of hidden states, as we empirically observe that it tends to increase during the training process. If not properly addressed, this increasing trend could lead to training instability. Considering the relationship $\mathrm{var}(\bm{a} + \bm{b}) = \mathrm{var}(\bm{a}) + \mathrm{var}(\bm{b})$.
We begin by analyzing the abnormal increase in variance of hidden states that arises inherently from residual connections.
Furthermore, we demonstrate how layer normalization can contribute to the variance of hidden states increase, particularly when the inputs deviate significantly from their normal range.
Finally, we show that certain abnormal updates in layer normalization are actually driven by the growing mean of attention logits.

\paragraph{Exploding hidden states due to residual connection}
Figure~\ref{fig:exploding_var} illustrates the exponential growth trend of the variance in the hidden states. To understand the underlying cause, we express the hidden states in terms of the model's weights and inputs:
\begin{align*}
    \mathrm{var}(\bm{z}^l) &= \mathrm{var}(\bm{y}^l) + \mathrm{var}(\texttt{FFN}(\texttt{RMSNorm}(\bm{y}^l))), \\
    \mathrm{var}(\bm{y}^l) &= \mathrm{var}(\bm{x}^l) + \mathrm{var}( \texttt{MHA}( \texttt{RMSNorm}(\bm{x}^l))).
\end{align*} 
For ease of analysis, we first assume that:
\begin{equation}
    \bm{x},\bm{y}\sim\mathcal{N}(0,\sigma^{2}). \label{eq:xy_assumption}
\end{equation}
Under this assumption, we can obtain $\text{var}(\bm{u}) = \text{var}(\bm{v}) = 1$. 
In this case, we can express the variance as the following form:
\begin{equation}
    \mathrm{var}(\bm{z}^l) = \mathrm{var}(\bm{x}^l) + \mathrm{var}(\texttt{FFN}(\bm{u})) + \mathrm{var}(\texttt{MHA}(\bm{v})), \label{eq:var_ln_layer_by_layer}
\end{equation}
which means, the hidden states will grow by the variance of \texttt{MHA} and \texttt{FFN} in each layer:
\begin{align}
    \mathrm{var}(\texttt{head}_i(\bm{v})) &= \mathrm{var}(\mathrm{softmax}(\mathbf{Z})\mathbf{V})\cdot d_\text{model}\cdot\mathrm{var}(\mathbf{W}_v)<d_\text{model}\cdot\mathrm{var}(\mathbf{W}_v), \\
    \mathrm{var}(\texttt{FFN}) &= d_\text{ffn}\cdot d_\text{model}\cdot\mathrm{var}(\mathbf{W}_1)\cdot\mathrm{var}(\mathbf{W}_2), \label{eq:var-ffn}\\
    \mathrm{var}(\texttt{MHA}) &= \mathrm{var}(\texttt{head}(\bm{v}))\cdot d_\text{model}\cdot\mathrm{var}(\mathbf{W}_o)< d_\text{model}^2\cdot\mathrm{var}(\mathbf{W}_v)\cdot\mathrm{var}(\mathbf{W}_o),\label{eq:var-attention}
\end{align}
where $\mathbf{Z}$ denotes the scaled attention scores.  
The base dimensionality $d_\text{model}$ of LLMs are often large (\eg 1,920 in our model).
Additionally, the vanilla setup in Hugging Face uses a default initialization standard deviation of 0.02 for the weight matrices. 
When training the proxy model, this initialization leads to significantly large variance values, exacerbated by the effects of the aforementioned \texttt{Attention} and \texttt{FFN} modules. 
As a result, the gradient norm becomes very large. A detailed derivation of the attention head can be found in~\cite{takase_spike_2024}. Common mitigation strategies include initializing the weights with a very small variance, typically inversely proportional to $d_\text{model}$, which we will discuss in Section~\ref{sec:stability_initialization}.

\paragraph{Exploding hidden states due to layer normalization}\label{sec:explosion-layer-normalization}

\begin{figure}[t]
    \centering
    \begin{minipage}{0.485\textwidth}
        \includegraphics[width=0.99\linewidth]{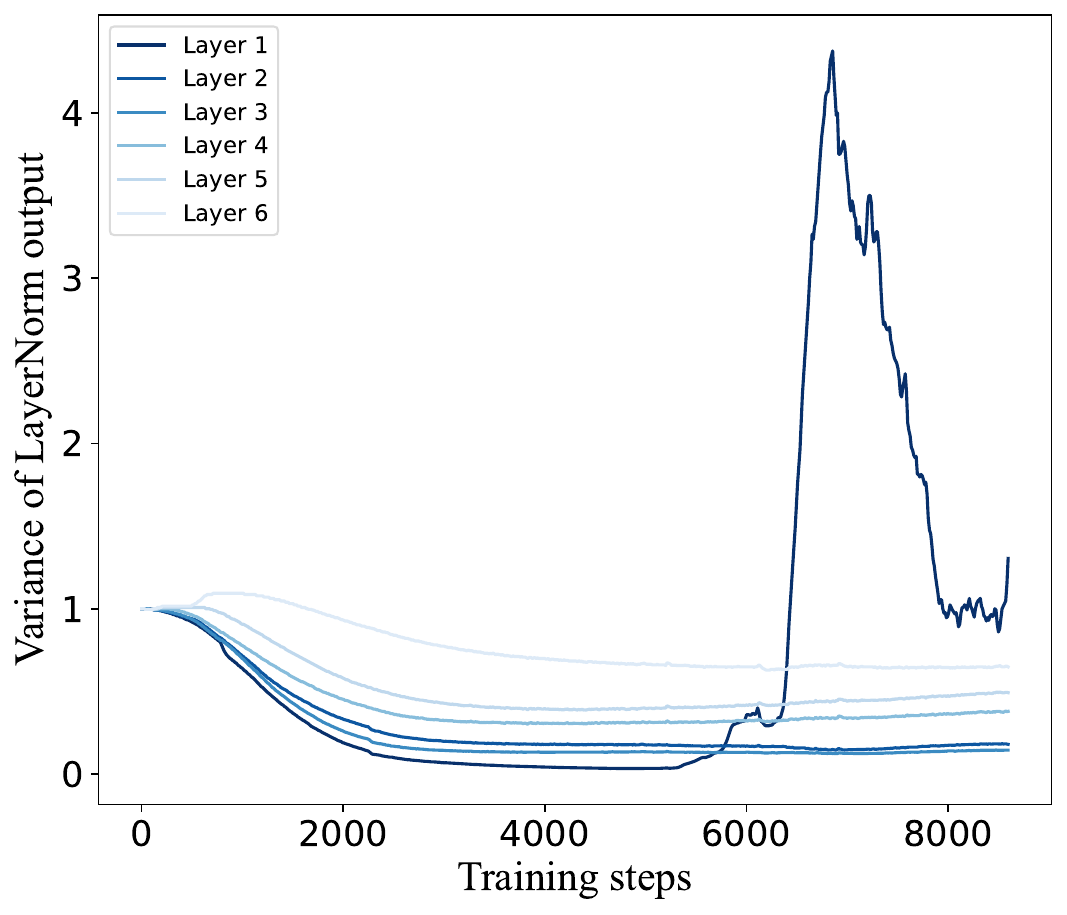}
        \caption{Variance of LN output of each layers.}
        \label{fig:exploding_hs_after_ln}
    \end{minipage}
    \hfill
    \begin{minipage}{0.485\textwidth}
        \includegraphics[width=0.99\linewidth]{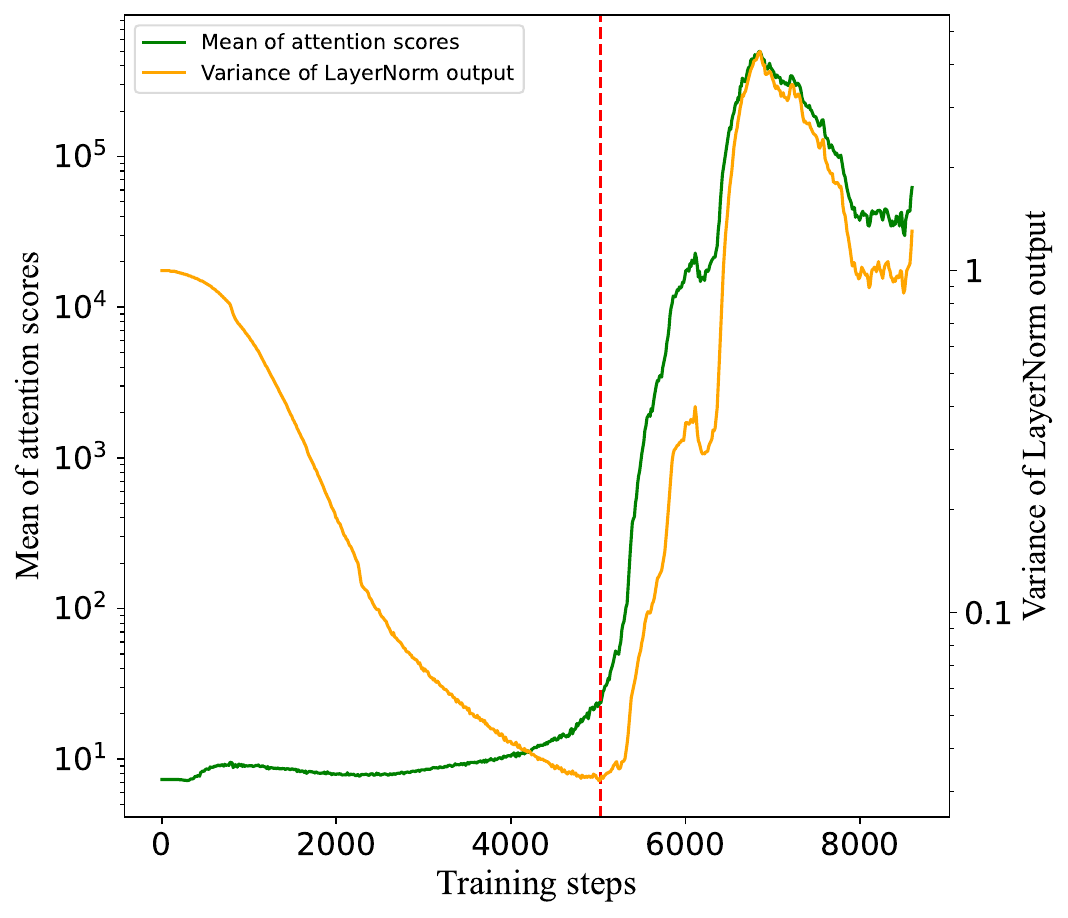}
        \caption{Attention scores explodes before LN.}
        \label{fig:exploding_attention_logits}
    \end{minipage}
\label{fig:exploding_attention_logits_main}
\end{figure}


According to prior studies~\citep{xiong_layer_2020,takase_spike_2024,wortsman_small-scale_2023}, the variance of the input to layer normalization should not be significantly smaller than 1, as this can lead to an increase in the gradient norm: 
$$
\left\Vert\frac{\partial \texttt{RMSNorm}(\bm{x})}{\partial \bm{x}}\right\Vert_2=\mathcal{O}\left(\frac{\sqrt{d}}{\Vert \bm{x}\Vert_2}\right).
$$
To investigate the impact of layer normalization, we conduct an experiment examining the variance of the outputs from the LN layer in our model (Figure~\ref{fig:exploding_hs_after_ln}), where the embedding layer is followed by layer normalization.
We empirically find, if the embedding layer is initialized with the default variance, we must multiply its output by a scaling factor to ensure the input to the layer normalization has a variance of 1. Therefore, we apply a scaling factor of $\texttt{scale\_embed}=10$ for model training. For experiments that do not employ embedding tying, we directly set the initialization standard deviation of the embeddings to 1.

\paragraph{Exploding hidden states due to attention scores}\label{sec:explosion-attention-logits}
When applying the scaled embedding method mentioned above, we have empirically observed that attention scores can also explode, resulting in the explosions of hidden states. We first  examine the calculation process of attention scores, which is formally defined as:
$$
\mathbf{S}=\mathbf{X}^T\mathbf{W}_Q^T\mathbf{W}_K\mathbf{X}. 
$$
Then, we calculate the gradient of the attention scores with respect to the query and key matrices as follows:
\begin{align}
    \frac{\partial \mathbf{S}}{\partial \mathbf{W}_Q} &= \mathbf{X} \mathbf{X}^T \mathbf{W}_K^T, \\
    \frac{\partial \mathbf{S}}{\partial \mathbf{W}_K} &= \mathbf{W}_Q^T \mathbf{X} \mathbf{X}^T. 
\end{align}
From the above derivation, we find that the query-key multiplication term involved can lead to an unbounded gradient, potentially causing severe self-excitation. In such cases, the mean of the attention scores gradually increases as the number of layers increases, eventually resulting in exploding hidden states. Therefore, it is essential to monitor and regularize the attention scores during training.

\subsection{Training Instability Mitigation Methods}\label{sec:stability_initialization}

After discussing the potential causes of training instability, we will introduce mitigation methods to enhance training stability.


\begin{figure}[t]
    \centering
    \includegraphics[width=0.8\textwidth]{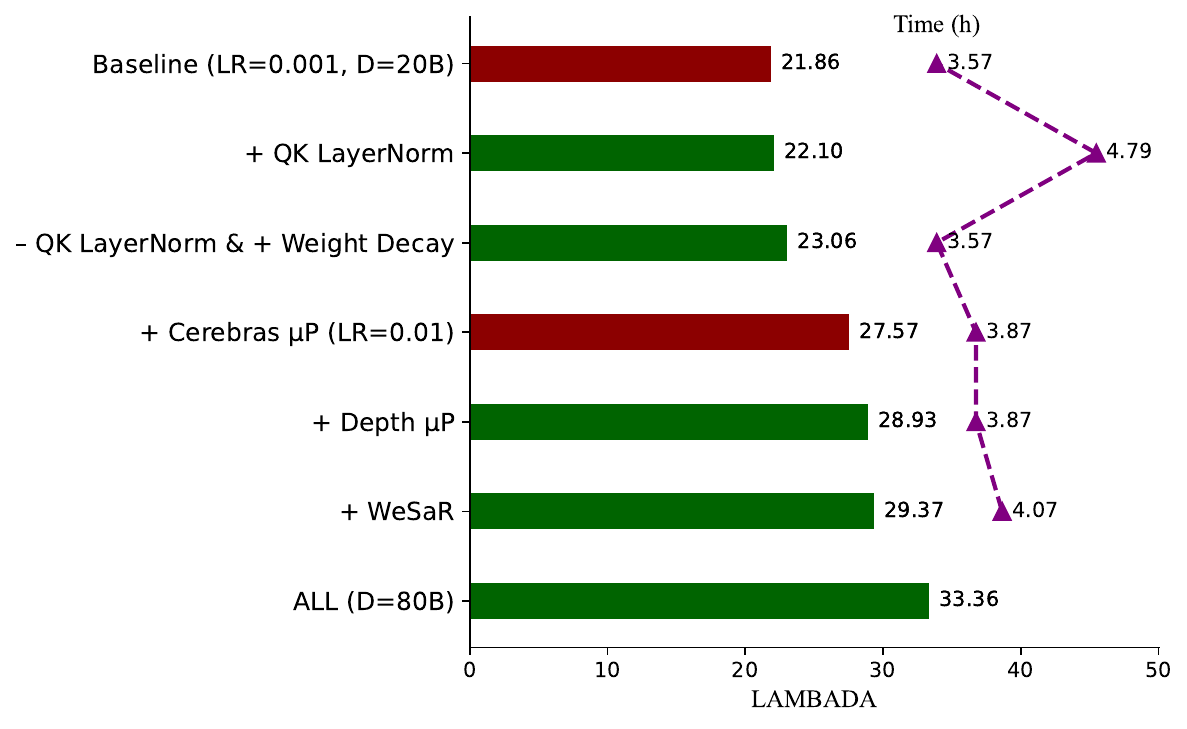}
    \caption{Ablation experiments on training instability mitigation methods are conducted. We report the average of LAMBADA accuracy of the last three checkpoints of the training and the estimated running time on our 48 A800-GPU cluster. Divergent gradient norm or spiking loss trajectories are shown in \textcolor{darkred}{red} bars, and convergent training is shown in \textcolor{darkgreen}{green}. }
    \label{fig:stability_experiments}
\end{figure}

\subsubsection{Scaled Initialization and Scaling Factor}\label{sec:scaled-init}

As discussed in Section~\ref{sec:hidden_states_theoretical_analysis}, employing a carefully designed initialization method along with appropriate scaling factors is a key strategy for addressing training instability. 
In this regard, we explore the initialization strategies proposed by Megatron-LM~\citep{shoeybi2020megatronlmtrainingmultibillionparameter} and BLOOM~\citep{scao_what_2022}. Specifically, we initialize $\mathbf{W}_2$  
of the \texttt{FFN} layer (Equation~(\ref{eq:ffn})) and $\mathbf{W}_o$ of the \texttt{Attention} layer (Equation~(\ref{eq:mha})) in accordance with $\mathcal{N}(0,\sigma_\text{base}^2/(2n_\text{layers}))$, and initialize the remaining parameters according to $\mathcal{N}(0,\sigma_\text{base}^2)$. Here, we set the initialization standard deviation as $\sigma_\text{base}=\sqrt{2/(5d)}$.

To analyze the effect of the above initialization strategy, we plug the initialization standard deviation into Equation~(\ref{eq:var-ffn}) and~(\ref{eq:var-attention}), and obtain the following formulas:
\begin{align*}
    \mathrm{var}(\texttt{FFN})  &=  r_\text{ffn}\cdot d_\text{model}^2\cdot \frac{2}{5 d_\text{model}} \cdot \frac{1}{5 d_\text{model} \cdot n_\text{layers}} =\frac{1}{5n_\text{layers}}, \\
    \mathrm{var}(\texttt{MHA}) &=\mathrm{var}(\texttt{head}(\bm{v}))\cdot d_\text{model}\cdot\frac{1}{5 d_\text{model} \cdot n_\text{layers}}  < \frac{2}{25n_\text{layers}}.
\end{align*}
Substituting the obtained results into Equation~(\ref{eq:var_ln_layer_by_layer}), we can measure the growth of the hidden states' variance throughout the entire network as:
\begin{equation}\label{var-bound}
\text{var}(\bm{z}) - \text{var}(\bm{x}) = n_\text{layers} \text{var}(\texttt{FFN}) + n_\text{layers} \text{var}(\texttt{MHA}) < 7/25,
\end{equation}
which successfully regularizes the growing trend of hidden states. 

{In the modern LLaMA architecture, it is more common to use a GLU-style of non-linearity, which can be denoted as $\texttt{FFN}(\bm{u})=[\mathcal{F}(\bm{u}\mathbf{W}_\text{gate})\odot(\bm{u}\mathbf{W}_\text{up})]\cdot\mathbf{W}_\text{down}$. We initialize $\mathbf{W}_\text{up}$ and $\mathbf{W}_\text{gate}$ the same way as $\mathbf{W}_1$, and $\mathbf{W}_\text{down}$ same as $\mathbf{W}_2$. The above derivation in Equation~(\ref{var-bound}) still holds empirically in this setting. 

By initializing in this manner, we  empirically find that while there may still be a tendency for the hidden states at each layer to gradually increase, this increase remains within a reasonable range.


\subsubsection{Maximal Update Parametrization}


In the above, we have explored the importance and effectiveness of parameter initialization through theoretical analysis and empirical experiments. However, when we migrate the training configuration selected on the 0.05B proxy model to the model of the target size, instability still exists. 


To ensure hyper-parameter consistency across different model scales, the Maximal Update Parametrization ($\mu$P) has been proposed~\citep{yang_tensor_2022,yang_tensor_2023}, including width scaling and depth scaling,
which facilitate transferring the hyperparameters from smaller models to the training of larger models. 
Typical use of this strategy can be found in CerebrasGPT 
~\citep{CerebrasGPT} and MiniCPM
~\citep{hu_minicpm_2024}. In particular, they have found that the optimal learning rate remains quite stable during migration. 
Compared to the basic method in Section~\ref{sec:scaled-init}, $\mu$P provides a more systematic approach for setting the initialization and scaling factor. 

We apply $\mu$P, which takes into account embedding parameters, model depth, and model width, for parameter initialization. Additionally, we conduct a comprehensive parameter search on proxy model to identify the optimal configuration, exploring parameters such as batch size and learning rate.

\begin{table*}[t]
    \centering
    \small
    \caption{Comparison of the used hyperparameter settings for training stability, where the detailed explanation for the variables are in Table~\ref{tab:stability_recipe_explain}.
   We include SI~\citep{takase_spike_2024} for comparison, MiniCPM~\citep{hu_minicpm_2024}, CerebrasGPT~\citep{CerebrasGPT}. The definition of the symbols is available at Table~\ref{tab:stability_recipe_explain} .}
    \renewcommand{\arraystretch}{1.4}
    \begin{center}
    {
    \begin{tabular}{lcccccc}
    \toprule
    \textbf{Method} &\textbf{SI} & \textbf{MiniCPM} & \textbf{CerebrasGPT}  & \textbf{YuLan-Mini}\\  
    \midrule[0.5pt]
    Scale Embedding Output & 1 &  12 & 10 & 10  \\
    
    Scale MHA equation   & $1/\sqrt{d_{\text{head}} }$ & $1/\sqrt{d_{\text{head}} }$& $1/d_{\text{head}}$  & $1/\sqrt{d_{\text{head}} }$  \\
    
    Scale Residual Connection 
    &1&$\frac{\text{1.4}}{\sqrt{n_{\text{layers}}}}$&1 &
    $\frac{\text{1.4}}{\sqrt{n_{\text{layers}}                               }}$\\
    
    QKV Weights LR &$\eta_{\text{base}}$&${\eta_{\text{base}}}/{m_{\text{width}}}$&${\eta_{\text{base}}}/{m_{\text{width}}}$ & ${\eta_{\text{base}}}/{m_{\text{width}}}$\\
    
    QKV $\sigma$ Init & ${\sigma_{\text{base}}^2}$ &      ${\sigma_{\text{base}}^2}/{m_{\text{width}}}$ &  ${\sigma_{\text{base}}^2}/{m_{\text{width}}}$  & ${\sigma_{\text{base}}^2}/{m_{\text{width}}}$\\
    
    O Weights LR &$\eta_{\text{base}}$&${\eta_{\text{base}}}/{m_{\text{width}}}$&${\eta_{\text{base}}}/{m_{\text{width}}}$& ${\eta_{\text{base}}}/{m_{\text{width}}}$\\
    
    O $\sigma$ Init &$\frac{\sigma_{\text{base}}^2}{2n_{\text{layers}}}$&${\sigma_{\text{base}}^2}/{m_{\text{width}}}$&$\frac{\sigma_{\text{base}}^2}{2m_{\text{width}}\cdot n_{\text{layers}}}$ & $\frac{\sigma_{\text{base}}^2}{2m_{\text{width}}\cdot n_{\text{layers}}}$\\
    
    FFN1 Weights LR &$\eta_{\text{base}}$&${\eta_{\text{base}}}/{m_{\text{width}}}$&${\eta_{\text{base}}}/{m_{\text{width}}}$& ${\eta_{\text{base}}}/{m_{\text{width}}}$\\
    
    FFN1 $\sigma$ Init &${\sigma_{\text{base}}^2}$&${\sigma_{\text{base}}^2}/{m_{\text{width}}}$&${\sigma_{\text{base}}^2}/{m_{\text{width}}}$& ${\sigma_{\text{base}}^2}/{m_{\text{width}}}$\\
    
    FFN2 Weights LR &$\eta_{\text{base}}$&${\eta_{\text{base}}}/{m_{\text{width}}}$&${\eta_{\text{base}}}/{m_{\text{width}}}$& ${\eta_{\text{base}}}/{m_{\text{width}}}$\\
    
    FFN2 $\sigma$ Init &$\frac{\sigma_{\text{base}}^2}{2n_{\text{layers}}}$&${\sigma_{\text{base}}^2}/{m_{\text{width}}}$&$\frac{\sigma_{\text{base}}^2}{2m_{\text{width}}\cdot n_{\text{layers}}}$&$\frac{\sigma_{\text{base}}^2}{2m_{\text{width}}\cdot n_{\text{layers}}}$\\
    
    Scale Output logits &1&${1}/{m_{\text{width}}}$&${1}/{m_{\text{width}}}$&1\\
    \bottomrule
    \end{tabular}}
    \end{center}
    \label{tab:stability_recipe}
\end{table*}

\subsubsection{Mitigating Instability through Re-Parametrization}


When using $\mu$P, we find that spikes in loss still occur under large learning rates. We speculate this is because, although  $\mu$P alleviates instability during the initial stages of training, it may still deviate from a stable state during prolonged updates at large learning rates.


Therefore, we apply a simple yet effective method WeSaR proposed by~\cite{nishida_initialization_2024}, which empirically decouples the update of the gradient norm from the gradient direction. This is achieved by re-parametrizing the matrix weights $\mathbf{W}$ with an additional learnable parameter \(\alpha\in\mathbb{R}\) as: 
\[
\mathbf{W} = \alpha \widetilde{\mathbf{W}}. 
\]

We find the above WeSaR method to be effective in addressing the reasons of exploding hidden states highlighted in the previous analysis, as illustrated in Figure~\ref{fig:convergent_var}.
It is likely due to re-parametrization, which distributes the gradient of a single weight across multiple new re-parametrized weights, thereby reducing the abnormal updates caused by excessively large gradients. 

Re-parametrization is applied to matrices other than Layer Normalization as shown in Figure~\ref{fig:stability_experiments}.
Specifically, we initialize $\widetilde{\mathbf{W}} \sim \mathcal{N}(0, \sigma^2)$ and set $\alpha = 1 / \gamma$. In this setting, it is straightforward to verify that $\mathbf{W} \sim \mathcal{N}(0, (\sigma / \gamma)^2)$, which satisfies the scaled initialization requirements described in Section~\ref{sec:scaled-init}.

\subsection{Discussion on Other Training Stabilization Methods}
During our training process, we thoroughly explore and utilize various training stabilization techniques. Below, we provide a brief introduction to these methods. 




\subsubsection{Warmup Based Methods}
To ensure the model transitions smoothly from its initial state to a stable training phase, we empirically find that employing learning rate warmup and sequence length warmup is often effective, which are detailed below. 

\paragraph{Learning rate warmup}
Learning rate warmup involves gradually increasing the learning rate from a small initial value (\eg $0$) to the max learning rate in $T_\text{LR}$ steps.
\cite{wortsman_small-scale_2023} suggests that a longer learning rate warmup can reduce sensitivity to the learning rate, as measured by training stability across different learning rates.
We empirically verify this conclusion and find increasing $T_{LR}$ indeed enhances training stability. For our final training, we set $T_\text{LR}=\,$2,433, which approximately corresponds to $10$ billion tokens of data.

\paragraph{Sequence length warmup}\label{sec:sequence_length_warmup}
Sequence length warmup starts training with short sequences (\eg $64$ tokens) and gradually increases their length within the steps of $T_{SL}$, which is typically set to a few multiples of $T_{LR}$~\citep{li2022stabilityefficiencydilemmainvestigatingsequence}. 
The rationale behind this approach is that longer sequence lengths contribute significantly to extreme gradient variance, particularly in the early stages of training. 
In our experiments, we also observe similar fluctuations in loss during long context training (especially in the 27-th curriculum phase). However, since we have stabilized the training using other methods and this approach requires additional preparation of the data, we ultimately decided not to adopt it.


\subsubsection{Module Based Methods}
In this part, we introduce module-based methods which regularize the model states by adjusting specific components in it. 

\paragraph{QK LayerNorm} 

QK LayerNorm and its variants (\eg QKV LayerNorm or capped QK LayerNorm) have have been shown to effectively mitigate the growth of attention logits~\citep{rybakov_methods_2024}, {which we also have identified in Section~\ref{sec:explosion-attention-logits}}. 
We highlight the effectiveness of QK LayerNorm because it directly addresses the exponential growth of gradients caused by the interaction of hidden states  ($\mathbf{Q}\mathbf{K}^T$), whereas some other methods only attempt to control the downstream instability.  
Our empirical study, which is shown in Figure~\ref{fig:qkln_attn_var} and~\ref{fig:qkln_grad_norm}, demonstrates the advantages of QK LayerNorm in terms of training stability. However, it significantly slows down the calculation in training: with the same acceleration configuration, using QK LayerNorm increases the training time by 34\%. 
Considering that the previously mentioned methods have already demonstrated stability in our preliminary experiments, we ultimately decided not to use QK LayerNorm.

\begin{figure}[t]
    \centering
    \begin{subfigure}[b]{0.485\textwidth}
        \centering
        \includegraphics[width=0.99\textwidth]{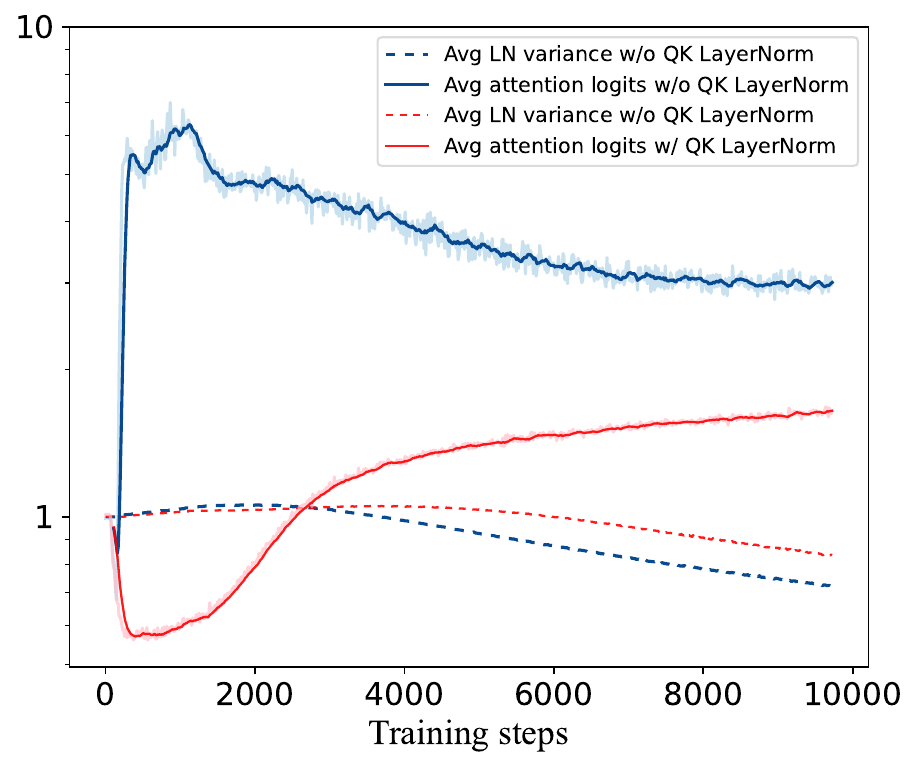}
        \caption{Variance of attention values and LN outputs}
        \label{fig:qkln_attn_var}
    \end{subfigure}
    \hspace{0.01\textwidth}
    \begin{subfigure}[b]{0.485\textwidth}
        \centering
        \includegraphics[width=0.99\textwidth]{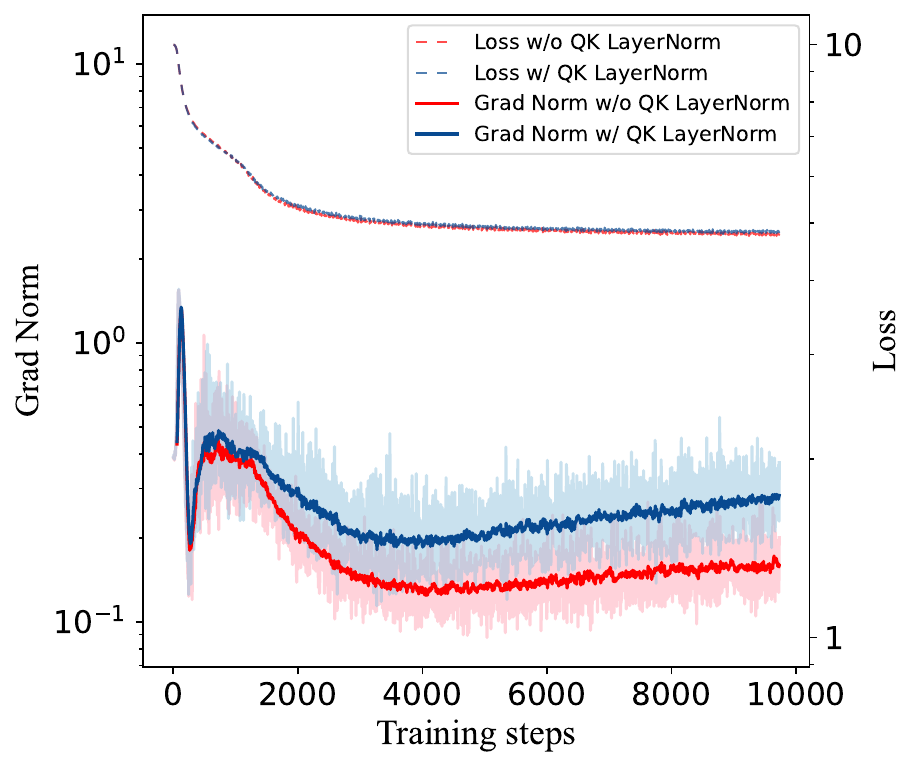}
        \caption{Gradient norm and loss trajectory}
        \label{fig:qkln_grad_norm}
    \end{subfigure}
    \caption{The curves of attention value and LN output variances (left) and gradient norm and loss (right). After using QK LayerNorm, we prevent the explosion of attention logits and gradients, keeping the LN output stable around 1 and the loss consistent.}

\label{fig:syn_compare}
\end{figure}

\paragraph{Embedding tying}
Embedding tying aims to share the weights of embedding and unembedding (\ie \texttt{lm\_head}) parameters~\citep{press-wolf-2017-using}. 
Our experiments demonstrate that the utilization of embedding sharing enables faster convergence and more stable training, and there is no significant degradation in training performance.


\paragraph{Z-loss}
Z-loss was originally proposed to alleviate the shift and scale of logits in classification tasks~\citep{brebisson_z-loss_2016}. Subsequently, it has been introduced to LLM and MoE training to mitigate the growth of the logits layer~\citep{palm-report,zoph2022stmoedesigningstabletransferable}. It adds an  auxiliary term related to the softmax normalizer $\log Z$ to the original loss: 
$\mathcal{L}=\texttt{lm\_loss} + \zeta\log^2 Z$.
In our experiments, we set the coefficient $\zeta=10^{-4}$ to encourage the logits to be close to $0$. Although ablation studies did not show significant effects, we incorporate it into the final training.


\subsubsection{Numerical Optimization Based Methods}
In addition, we consider using several commons methods to  reduce abnormal updates during optimization, as described below. 



\paragraph{Weight decay}

To prevent abnormal model weights due to large gradient updates, weight decay functions by subtracting a penalty term from the weights during the update step, rather than directly modifying the gradients. 
Formally, we denote the AdamW update without learning rate or weight decay as:
\begin{equation}
    \Delta = \alpha\hat{\bm{m}}_t/(\sqrt{\hat{\bm{v}}_t} + \epsilon). \label{eq:adamw-update}
\end{equation}
Then at update step $t$, the AdamW update with weight decay is given by  $\theta\rightarrow\theta-s_t\eta(\Delta - \lambda\theta)$, where $\lambda$ is the weight decay coefficient, $s_t$ is learning rate schedule and $\eta$ is the max learning rate. Previous work has recommended using an independent weight decay for updates, expressed as $\theta\rightarrow\theta-s_t(\eta\Delta - \lambda^\prime\theta)$, which is claimed to be applicable to a wider range of learning rates~\citep{loshchilovDecoupledWeightDecay2019,wortsman_small-scale_2023}. In the PyTorch implementation, this approach can be achieved by tuning the weight decay coefficient $\lambda$ 
in conjunction with the maximum learning rate, following the relationship $\lambda^\prime=\eta\cdot\lambda$.




\paragraph{Optimizer hyper-parameter}
In the update of AdamW (Equation~(\ref{eq:adamw-update})), $\hat{\bm{m}}_t$ and $\hat{\bm{v}}_t$ represent the first and second gradient moment exponential moving averages (EMA), respectively. If the gradient is of the same order of magnitude as $\epsilon$, then the update value $\Delta$ will be significantly reduced due to $\epsilon$, which empirically leads to training instability inherent in embedding layer.
A direct solution is to reduce $\epsilon$ from the default value of $10^{-8}$ to $10^{-15}$. Generally speaking, this method can alleviate the divergence caused by abnormal embedding gradient values in larger-scale models~\citep{wortsman_small-scale_2023,molybog_theory_2023}.



\paragraph{Numerical stability} In practice, paying close attention to numerical stability is crucial, as it can be an important source of training instability. In large-scale model training, \texttt{float32} often suffers from low computational efficiency. Although \texttt{float16} offers comparable precision with higher computational efficiency, it has a limited numerical representation range (\eg maximum positive number that can be represented is 65,504). Therefore, \texttt{bfloat16} has been proposed as a trade-off between precision and representation range. It largely alleviates the training instability caused by exceeding the representable range.
However, in practice, \texttt{bfloat16} introduces precision problems compared to \texttt{float16}. In experiments conducted by \cite{lee_fp8_2024} using \texttt{bfloat16} with 188 random seeds, 18 runs diverged, whereas using \texttt{float32} under the same configuration resulted in all runs converging normally. To mitigate precision issues with \texttt{bfloat16}, Gemma~\citep{gemma-report} find that shifting the RMSNorm weight from 1 to 0 helps, considering that \texttt{bfloat16} has symmetric numerical precision around 0 but greater inaccuracies near 1.

\paragraph{Value clipping}
To further limit the gradient within certain range, we utilize a gradient clipping of $1$. We find using a smaller limit does not help stabilize the training. In addition, initializing the LLM in accordance with ``$3$-$\sigma$'' rule with \texttt{nn.init.trunc\_normal\_} may be helpful for numerical stability.
\section{Data Pipeline}\label{sec:data_pipeline}

To pre-train an effective LLM, it is crucial to develop a robust data pipeline that comprehensively encompasses the key steps for  curating the pre-training data. These steps include data collection, filtering, selection, mixing, and curriculum design. Below, we describe each step of the data pipeline in detail.



\subsection{Data Collection}


For pre-training data preparation, we primarily reference the data configuration of Yulan-3~\citep{zhu_yulan_2024} and Llama-3-SynE~\citep{llama-3-syne}, which encompasses a wide-ranging and diverse collection of data such as web pages, encyclopedias, books, mathematical corpora, code, general knowledge, and synthetic data. Table~\ref{tab:data_stat} presents an overall summary regarding the composition of our training data. 


\begin{table}[!t]
\centering
\small
\caption{Statistical information of the entire pre-training corpus for YuLan-Mini. The data during the annealing process is detailed in Table~\ref{tab:decay_data_stat}. For model reproducibility, all curated datasets are placed in Appendix~\ref{sec:data-list}, and the remaining synthetic data we generated is open-sourced. }
\begin{center}
\begin{tabular}{llr}
    \toprule
        \textbf{Type} & \textbf{Source}   &  \textbf{Volume} \\
    \midrule
        Web Pages & FineWeb-Edu, DCLM, Chinese-FineWeb-Edu  & 559.76B \\
        Math (Pretrain) & AutoMathText, Proof-Pile-2, OpenWebMath Pro  & 85.00B \\
        Code (Pretrain) & the-stack-v2, StarCoder  & 202.44B \\
        General Knowledge & arXiv, StackExchange, English News  & 121.87B \\
        Books & CBook, Gutenberg, LoC-PD-Books & 52.13B \\
        Encyclopedia & Wikipedia, Baidu-Baike  & 14.80B \\
        Open-Source Instruction  & SlimOrca,  OpenMathInstruct-1, JiuZhang3.0  & 11.64B \\
        Synthetic Pretrain Data (Ours)  & Synthetic document (seed: AutoMathText, LeetCode)  & 8.76B \\
        Synthetic Instruction (Ours)  &  Reasoning (seed: MetaMathQA, DeepMind Math, ...)  & 23.52B \\
    \midrule
        Total &  - & 1,080B \\
    \bottomrule
\end{tabular}
\end{center}
\label{tab:data_stat}
\end{table}







\subsection{Data Filtering}\label{sec:data-prun}


As we aim for a data-efficient training approach, data quality is crucial to the final model's performance. For this purpose, we implement a thorough data cleaning process to remove low-quality texts (Figure~\ref{fig:data_filtering}).



\paragraph{De-duplication}
Data de-duplication is a crucial step in standard LLM training practices, as previous research has demonstrated that duplicate data can significantly degrade model performance~\citep{data-dedup}. We use the MinHash algorithm implemented by the Yulan-GARDEN library~\citep{yulan-garden} to deduplicate the training data.

\paragraph{Heuristic filtering} We adopt heuristic methods  to filter the data, some of which are listed as follows: 

\begin{itemize}
    \item All: we remove the documents containing fewer than 20 tokens.
    \item Code: we apply filtering criteria based on code metrics (\eg average line length, alphabetic characters ratio, and keyword statistics) similar to DeepSeek-Coder~\citep{deepseek-coder}.
    \item Synthetic data: we remove responses  that are garbled or contain repeated content. For math texts, we remove response that do not contain an hightlited answer part (\eg \texttt{\$box\{\}\$}). 
\end{itemize}


\paragraph{Topic-based text recall}

To enhance the model's capabilities in specialized areas, it is essential to include ample knowledge documents related to mathematics, code, and reasoning. 
For this purpose, we extract relevant documents from unused web pages by training \texttt{fasttext}~\citep{bojanowski2017enriching} and TinyBert~\citep{jiao_tinybert_2020} classifiers specifically tailored to these categories. From the {FineWeb-Edu}~\citep{lozhkov2024fineweb-edu} and {DCLM}~\citep{dclm-report} web corpus, we extract 10.4B math text tokens, 1.11B code text tokens, and 1.01B reasoning text tokens. which are directly used for training or serve as seed data for synthesizing instruction data. 
Furthermore, we reuse the synthesized science data (1.5B) from Llama-3-SynE~\citep{llama-3-syne}, which covers an extensive range of disciplines, such as math and physics. 



\paragraph{Model-based quality scoring}
For general web page data and mathematical pre-training data, we use the \texttt{fineweb-edu-scorer} released by FineWeb-Edu for data scoring. For Python code data, we use the \texttt{python-edu-scorer} released by FineWeb-Edu. To avoid language models favoring highly technical pages like arXiv abstracts and submitted papers, these two classifiers focus on knowledge at the elementary and middle school levels. 
Following the methodology of \cite{penedo_fineweb_2024}, we conduct quality assessments on all Python code data, most mathematical data, and web page data using scoring tools. We exclude data with scores of 1 and 2 and then heuristically sort data with scores from 3 to 5 (as detailed in Section \ref{sec:data-curr}). 

\paragraph{Decontamination}
To ensure the fairness of comparison, we perform decontamination based on the selected evaluation benchmarks. 
Initially, we tokenize both the training set and the benchmarks that require decontamination, such as GSM8K~\citep{gsm8k}, MATH~\citep{math}, HumanEval~\citep{humaneval}, and ARC~\citep{arc}.
Next, we divide all the benchmarks using $n$-gram tokens to create a contamination set. We use tokens rather than words to form $n$-gram segment, which achieves a higher level of decontamination in the domains of mathematics and code. 
Additionally, we exclude 20-gram segments that occur more than four times, as they are typically not relevant to the questions or solutions. Ultimately, the contamination set comprises 1,917,428 tuples. For each training document, if more than 10\% of its generated 20-grams are present in the contamination set, we exclude that document from the final pre-training set.


\begin{figure}
    \centering
    \includegraphics[width=\linewidth]{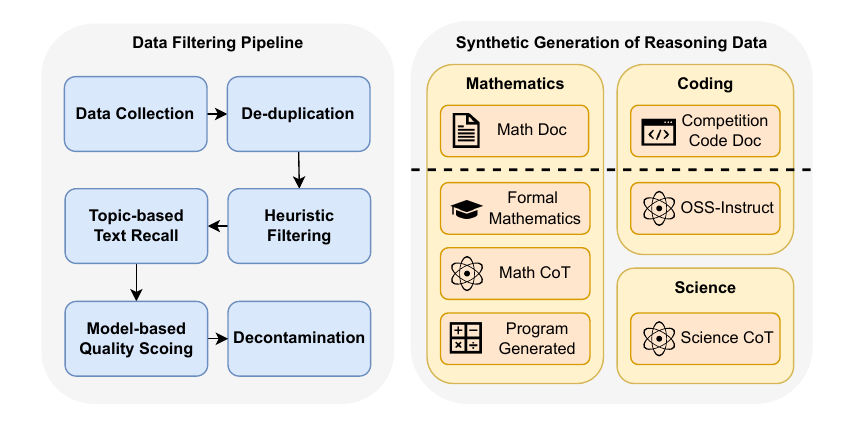}
    \caption{Illustration of our data filtering pipeline and synthetic generation for reasoning data. The filtering pipeline consists of six steps starting from data collection. Synthetic data generation includes both pretraining data (above the horizontal line) and instruction data (below the line).}
    \label{fig:data_filtering}
\end{figure}

\subsection{Synthetic Generation of Reasoning Data}\label{sec:synthetic-data}

Reasoning is widely recognized as one of the most desirable capabilities for LLMs~\citep{huang-chang-2023-towards}. However, unlike basic skills such as knowledge memorization, reasoning is more challenging to achieve and enhance in LLMs. One potential reason for this difficulty is the scarcity of high-quality texts containing logical or complex reasoning in real-world datasets. To address this limitation, a common approach is to generate data samples related to reasoning, such as chain-of-thought data, which focus on demonstrating the thought processes likely to lead to the final solution rather than directly producing question-answer pairs~\citep{chain-of-thought}. To enrich diversity and coverage, we consider a broad range of domains for synthetic data generation, as introduced below.

\subsubsection{Math Reasoning}
Mathematical reasoning data is relatively scarce in pre-training corpora, yet it is crucial for enhancing model capabilities~\citep{chain-of-thought}. To address this gap, we generate a diverse range of mathematical reasoning data, including documents, instructions, and formal mathematics data.

\paragraph{Mathematical documents} 
For mathematical documents, we generate descriptive content spanning various difficulty levels and thematic styles. It includes explanations of mathematical concepts and science education materials for primary school levels, as well as lecture scripts, tutorials, educational articles, and problem sets suitable for high school and college-level content. 
We primarily source math-related seed data from mathematical pre-training corpora, such as OpenWebMath~\citep{openwebmath}, and from a self-compiled math dataset using classifiers described in Section~\ref{sec:data-prun}.

\paragraph{Chain-of-thought reasoning} For instructional data, we employ three approaches for text synthesis. First, we use \texttt{Qwen2.5-Math-7B-Instruct}~\citep{qwen2.5-math} to generate thought processes (\eg chain-of-thought) for existing problems found in open-source datasets, such as Orca-Math~\citep{orca-math}, MetaMathQA~\citep{metamathqa}, AMPS-Math~\citep{math}, and NuminaMath~\citep{numina_math_datasets}. Second, following the method of JiuZhang3.0~\citep{jiuzhang3.0}, we utilize a finetuned \texttt{Qwen2-Math-7B-Instruct}~\citep{ding_unleashing_2024} to automatically generate new math problems (without the thought process) and then annotate the solutions in the same manner as the first approach. To obtain more extensive thought processes, we select more challenging data, such as from the NuminaMath dataset, and utilize slow-thinking model  \texttt{QwQ-32B-Preview}~\citep{qwq-32b-preview} 
for distillation. We obtain   the long-form thought data from our o1-reproduction project ``\emph{Slow thinking with LLMs}''~\citep{Slow_Thinking_with_LLMs_1,Slow_Thinking_with_LLMs_2}. This data aims to enhance the mathematical capacities of our base model. 

\paragraph{Formal mathematical reasoning} We also incorporate reasoning data from formal mathematics, such as formal theorem proving, which has been shown to enhance performance on general mathematical tests like GSM8K and MATH. Specifically, we collect Lean tactic (\eg\texttt{intro}, \texttt{simp}) dataset like DeepSeek-Prover~\citep{xin_deepseek-prover_2024} and its associated prover states to train the model in generating proof tactics.
For the Lean GitHub dataset~\citep{wu_lean-github_2024}, we concatenate together the reasoning steps in each document to form a long reasoning chain.
Additionally, inspired by LIME~\citep{wu_lime_2022}, we augment the Lean Workbook~\citep{ying_lean_2024} datasets with three reasoning primitives: (1) Deduction: $\text{State}_\text{before}, \text{Tactic} \rightarrow \text{State}_\text{after}$; (2) Abduction: $\text{State}_\text{after}, \text{Tactic} \rightarrow \text{State}_\text{before}$; and (3) Induction: $\text{State}_\text{before}, \text{State}_\text{after} \rightarrow \text{Tactic}$. Rather than directly predicting the next proof tactics, we train the model to predict the previous or next state based on the proof tactics.

\paragraph{Program generated numerical reasoning}

LLMs perform reasoning in natural language, which is flexible but does not allow for the verification of the correctness and necessity of each step. To enhance the model’s basic numerical abilities, we select subsets of addition, subtraction, multiplication, division, and remainder operations from the DeepMind-Math dataset~\citep{deepmind-math}. Our goal is to transform simple mathematical expressions (\eg ``\texttt{What is 0.079 - 162?}'') into a corresponding calculation procedure consisting of individual steps. This enables the model to learn calculations in a manner similar to chain-of-thought reasoning. This approach is mainly applicable to simple and limited mathematical computations and is suitable only for the early and middle stages of training. Given that manually writing conversion code is time-consuming, we also utilize an agentic framework for the automatic generation of code.

\subsubsection{Code Reasoning}
For code reasoning, we primarily synthesize two types of data: programming competition problems and real-world programming tasks. We also generate long-form reasoning thought data with slow-thinking models.

\paragraph{Competition code synthesis through ICL} 
To enhance existing programming competition datasets, like those from LeetCode\footnote{\url{https://huggingface.co/datasets/greengerong/leetcode}}, we utilize the in-context learning (ICL) method by leveraging a small number of demonstrations. This approach allows us to generate additional examples, thereby expanding and diversifying the data. By introducing a broader range of challenging programming problems, we enrich the dataset significantly.


\paragraph{OSS-Instruct} Additionally, we generate real-world programming tasks and their corresponding solutions using the OSS-Instruct method, as detailed in previous work~\citep{magicoder}. This process is guided by carefully crafted prompts (see Appendix~\ref{sec:appendix-prompt}), ensuring that the generated tasks are both relevant and applicable to real-world scenarios. This approach significantly enhances the practical utility of the dataset.

  


\subsubsection{Scientific Reasoning}
Scientific reasoning is crucial for expanding the capabilities of LLMs. To acquire scientific reasoning data, we consider the following two approaches.

\paragraph{Scientific chain-of-thought reasoning} {In Section~\ref{sec:data-prun}, we train a scientific classifier to extract documents related to various scientific fields from the web page data of {FineWeb-Edu} and {DCLM}. We utilize these scientific data as the seed data, and apply a synthesis method similar to that used for generating math reasoning data to create science-related questions and answers~\citep{llama-3-syne}. 
}




\paragraph{Scientific problems with slow-thinking processes} We gather more difficult scientific questions from college entrance examinations and camel-ai, covering subjects such as physics, chemistry, and biology. Then we use \texttt{QwQ-32B-Preview} 
to answer questions and obtain pairs of answers to difficult scientific data questions.

\subsubsection{Reflection}
An important aspect of reasoning ability is the capacity to reflect on and backtrack from the current state~\citep{reflexion}. In this work, we explore the generation of reflection data to further enhance the model's reasoning capabilities. 
First, we sample mathematical problems and collect both positive and negative responses by comparing them to the golden label. We then employ a powerful model, \texttt{Qwen2.5-Math-7B-Instruct}, to identify the first error in the negative response and truncate the content that follows. Instead of merely concatenating the positive response with the truncated negative one, we create error analyses and transitional statements to seamlessly and effectively connect the two responses.


The prompts used for data synthesis are provided in Appendix~\ref{sec:appendix-prompt}. 


\subsection{Data Mix}\label{sec:data-mix}

Our training process is divided into three main stages: warmup, stable training, and annealing. 

During the warmup and stable training stages, the dataset composition is as follows: 60\% general English data (consisting of 45\% from web and 15\% from books, papers and other relevant sources), 20\% code data, 10\% math data, and 10\% general Chinese data. In the later stage of  stable training, a small amount (<5\%) of instruction data is introduced. We maintain a relatively consistent data distribution across different curriculum phases, and will slightly adjust the data proportions according to the the perplexity performance of the model on  various benchmarks. 
We strive to avoid large shifts in data distribution, as significant changes can cause the loss to spike suddenly. To ensure training stability and account for testing inaccuracies, the change in data distribution between two consecutive phases is kept within 3\%.


During the annealing stage, the proportion of instruction data is increased to 19.19\% in total: code-related instruction data constitutes approximately 11\%, math-related instruction data accounts for about 7\%, and general instruction data amounts to around 1\%. Additionally, the proportion of long context data (\ie those exceeding 8K tokens) is also increased, occupying approximately 14.21\% of the tokens used.

\subsection{Data Curriculum}\label{sec:data-curr}

Following previous studies~\citep{data-ordering,llama-3-syne}, we adopt a curriculum-based method to prepare training data throughout the process. We use quality classifiers, such as \texttt{fineweb-edu-scorer}, to evaluate content based on educational difficulty, assigning higher scores to text content suitable for primary and secondary school levels. 
Manual inspection reveals that, for math and code pre-training data, lower scores often indicate higher difficulty. Conversely, for English webpage data, we find that staged training based on educational level scores significantly impacts the original distribution. Therefore, we train on math and code data in order of increasing difficulty, but do not apply curriculum learning to webpage data.



Throughout the entire training process, we continuously monitor the model's performance, as detailed in Section~\ref{sec:monitor}. We save a checkpoint and conduct an evaluation each 4B training tokens. For each 40B tokens, we reassess and adjust the data ratio when transitioning between training phases based on the model's overall performance in that phase. For example, if the model's performance on the HumanEval benchmark does not improve or declines after a stage, we may consider slightly increasing the amount of code data in the subsequent stage.


Additionally, to help the model adapt to a relatively high proportion of instruction data in the annealing stage, we gradually increase the amount of instruction data. However, throughout the entire stable training stage, this proportion will not exceed 5\%.

We present the data distribution scheduled in the training  curriculum in Figure~\ref{fig:data_mix}, and the detailed data composition for each phase are presented in Appendix~\ref{data-all-data}. 

\begin{figure}
    \centering
    \includegraphics[width=1\linewidth]{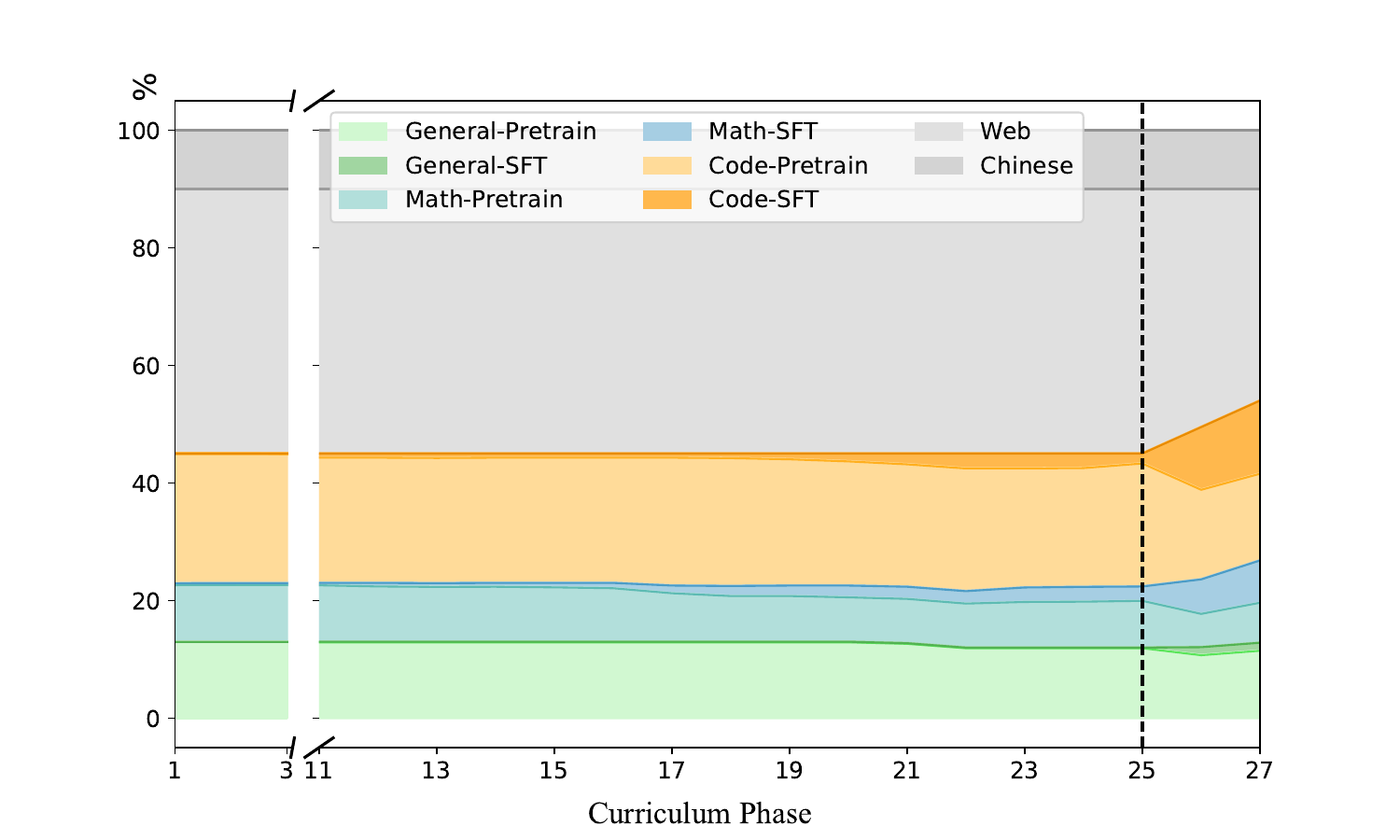}
    \caption{The data mixture proportion of \emph{math}, \emph{code}, and \emph{general} data. We keep the proportion of web data unchanged in stable stage, and then gradually decrease it in annealing stage.
    The entire process is divided into into three major stages: warmup, stable training, and annealing (\ie beginning after the dashed line). 
    }
    \label{fig:data_mix}
\end{figure}







\section{Annealing}\label{sec:decay}
As demonstrated in previous studies~\citep{hu_minicpm_2024}, the annealing stage is particularly effective in boosting model performance. We also implement effective training strategies during the annealing stage, which are described below.


\subsection{Optimization Setting}

In the annealing stage, we improve the performance of the model by using a high-quality data set and equip the model with long context processing capacities. Thus, we mainly consider the two settings, \ie learning rate annealing and context window extension. 
 
\paragraph{Learning rate annealing}\label{sec:lr-decay-f}Since we use the WSD scheduler, in the annealing stage, the learning rate will gradually decrease from that in the stable training stage. We need to select an appropriate learning rate annealing function. We investigate the performance of several typical learning rate annealing functions~\citep{DBLP:journals/corr/abs-2405-18392}, such as linear annealing, cosine annealing, 1-sqrt annealing. We empirically find  that 1-sqrt performs the best, so we choose 1-sqrt as our learning rate annealing function, which is defined as follows:
$$
f(n; N, N_{\text{annealing}}) = \left( 1 - \sqrt{\frac{n - (N - N_{\text{annealing}})}{N_{\text{annealing}}}} \right),
$$
where $n$, $N$ and $N_{\text{annealing}}$ denote the current number of  steps, the total number of steps and the number of annealing steps, respectively. 

Following the work by \cite{hu_minicpm_2024}, we estimate the optimal annealing ratio to be 8\%, \ie 80 billion tokens. We maintain the same batch size used during stable training, \ie 4 million tokens. The learning rate is decreased from $10^{-2}$ to $5.22 \times 10^{-5}$ over a span of 18,802 steps. 
Subsequently, the learning rate is held constant at $5.22 \times 10^{-5}$ for the final 772 steps.


\paragraph{Context Window Extension}\label{sec:context_window_extension} 
Previous research~\citep{chen-arxiv-2023-extending} has demonstrated that LLMs can hardly process texts exceeding their context windows due to the out-of-distribution (OOD) rotation angles in RoPE. To achieve the context window extension, increasing the base frequency of RoPE to migrate the OOD rotation angles and continual pre-training has been an effective method~\citep{xiong-etal-2024-effective}.
Consequently, during the annealing stage, we increase the base frequency of RoPE $\theta$ from 10,000, employed during stable training, to 490,000 and train the model on long texts.
This adjustment successfully extends the context length from 4,096 (4K) tokens to 28,672 (28K) tokens.






\subsection{Data Selection for Annealing Stage}
It is particularly important to select high-quality data during the annealing stage~\citep{hu_minicpm_2024}. As the learning rate decreases, the model can rapidly enhance its performance by leveraging high-quality data. We consider selecting high-quality data for the annealing stage using the following methods:

\begin{itemize}
\item We include a variety of high-quality data sources, particularly synthetic reasoning data discussed in Section~\ref{sec:synthetic-data}. 
\item We use a gradient-based data selection method, which accelerates and improves the LESS method~\citep{DBLP:conf/icml/XiaMGA024}, combining the method InsTag~\citep{DBLP:conf/iclr/LuY0LLTZZ24} for constructing a diversified target set.
\end{itemize}

In particular, we incorporate formal mathematical reasoning (theorem proving in Lean) and advanced reasoning data (o1-like thought data) to improve the model's performance on challenging math benchmarks, \eg MATH-500, which have been shown in Table~\ref{tab:math_code_benchmark}. 

We present the final data composition for the annealing stage in Table~\ref{tab:decay_data_stat}.

\begin{table}[!t]
\centering
\small
\caption{Detailed information of the training data in the annealing stage. }
\begin{center}
\begin{tabular}{lllr}
    \toprule
        \textbf{Domain} & \textbf{Type} & \textbf{Dataset} &  \textbf{Volume} \\
    \midrule
        Mix & Pretrain & FineWeb-Edu, CBook, arXiv & 64.65B \\
        Math & (1) CoT &Deepmind-Math,    MathInstruct  & {3.07B} \\
        & (2) Long CoT & {Numina, AMPS, Platypus} & 0.61B \\
        & (3) Formal math & {Lean-GitHub}, {Lean-WorkBook}, {DeepSeek-Prover-V1} & 0.10B \\
        & (4) Curated & {Tulu v3, MathInstruct} & 1.42B \\
        Code & (1) CoT &  {OSS-Instruct (seed: the-Stack-v2)}, {OpenCoder-LLM} & 6.66B\\
        
        
        & (2) Curated & {LeetCode, XCoder-80K} & 2.39B\\
        Science & (1) Long CoT & Camel-ai & 0.04B\\
        & (2) Curated & {EvolKit-20k, Celestia, Supernova} & 1.06B\\
    \midrule
        Total & - & - & 80B \\
    \bottomrule
\end{tabular}
\end{center}
\label{tab:decay_data_stat}
\end{table}

\subsection{Long Context Training}  

During the annealing stage of the final 80B tokens, we adjust the base frequency of RoPE from 10,000 to 490,000 and train on long sequences to extend the context length from 4,096 tokens to 28,672 tokens.
We avoid training with long contexts in earlier stages because the computational cost of self-attention layers increases quadratically with sequence length, making it prohibitively expensive~\citep{llama-3-card}. 

When training on long contexts, we observe a decline in the model's performance on short-text benchmarks. To enhance the long-text capacities and preserve the short-text capacities, we carefully design the mixture of data. We {upample books and concatenated GitHub code texts} ~\citep{lv_longwanjuan_2024} as long context data to capture long-term dependencies, while using high-quality short texts to preserve short-text capabilities.
Additionally, inspired by previous studies~\cite{ding2024fewertruncationsimprovelanguage,gao2024trainlongcontextlanguagemodels}, we also apply masked cross-document attention that prevents attention across different documents to preserve short-context capabilities. 



\subsection{Other Strategies}




\paragraph{Packing}

Since the training data during the annealing stage includes some instruction data, using a traditional simple {packing method} for pre-training data could result in instruction data being split, thereby compromising its effectiveness. To address this, we propose a packing strategy designed to maintain training efficiency while minimizing the disruption of instruction data. This strategy involves different packing methods based on data type. Pre-training data is directly spliced, whereas for instruction data, if it is divided into two sequences, the remaining part of the previous sequence is padded directly, and this instruction data serves as the beginning of the second sequence. Subsequently, any redundant padding tokens are replaced with pre-training data tokens.
By including the instruction data, our main goal is to learn the reasoning process rather than focusing solely on the question-and-answer format. Therefore, we employ the same data processing method used in pre-training, which directly includes question-answer pairs without relying on a chat template. When calculating the loss, the instruction and response are treated as a single document, and the loss for the instruction is not masked.



\paragraph{Checkpoint merging}

Following the approach used in LLaMA3 \citep{llama-3-card}, we combine the last few checkpoints during the annealing stage to produce the final pre-trained model. While this strategy might result in a slight reduction in certain specific capabilities (\eg GSM8K), it generally leads to a more well-rounded model.


\section{Evaluation}\label{sec:6}

In this section, we conduct the evaluation experiments to verify the effectiveness of our base model YuLan-Mini. We first set up the experiments for evaluation, and then present the experiment results. 

\subsection{Experimental Setup}
\subsubsection{Evaluation Benchmarks}

For a comprehensive evaluation of LLMs performance, we select the benchmarks from the following aspects.

\begin{itemize}
    \item \textit{Language comprehension}: We select the widely-used English benchmarks MMLU~\citep{mmlu}, LAMBADA~\citep{lambada} and RACE~\citep{lai2017racelargescalereadingcomprehension}, along with the Chinese benchmarks CMMLU~\citep{cmmlu} and CEval~\citep{huang2023ceval}, to evaluate the bilingual comprehension capabilities of the LLM. These benchmarks span various domains, such as history, science, and culture.

    \item \textit{Code generation}: We select Humaneval~\citep{humaneval} and MBPP~\citep{mbpp} to assess the capability of LLMs to generate accurate code snippets for natural language problems.

    \item \textit{Mathematical reasoning}: We utilize GSM8K~\citep{gsm8k} and MATH-500~\citep{math,verify} to evaluate the mathematical reasoning capabilities of LLMs. These benchmarks range from basic arithmetic to advanced mathematical problems.

    \item \textit{Logical reasoning}: We assess the logical reasoning capabilities of LLMs using 
    ARC-E~\citep{arc}, ARC-C~\citep{arc}, 
    which provide a comprehensive evaluation of logical reasoning across various knowledge domains.


    \item \textit{Commonsense reasoning}: We evaluate the LLM's commonsense reasoning ability using WinoGrande~\citep{winogrande}, HellaSwag~\citep{hellaswag}, StoryCloze~\citep{mostafazadeh2016corpusevaluationframeworkdeeper} which test the understanding and utilization of daily commonsense knowledge.
    
    \item \textit{Long context understanding}: We employ RULER~\citep{ruler} to evaluate the long context understanding ability, which measures its performance change as the sequence length increases. We perform evaluations of applicable models within a context length of 28K tokens. 

\end{itemize}



\subsubsection{Baseline Models}

To ensure a comprehensive evaluation, we select several small LLMs with comparable scales (\ie base models ranging from 0.5 to 3B, including embedding sizes) as baselines for comparison: 

\begin{itemize}
    \item \textit{MiniCPM-2.4B}~\citep{hu_minicpm_2024}: MiniCPM-2.4B {is pre-trained on 1.06T tokens and also employs the annealing training strategy}. Despite its small size, it exhibits impressive performance in general tasks while supporting deployments with limited hardware resource. 

    \item \textit{Qwen series models}~\citep{qwen2.5,qwen2}:
We select Qwen2-1.5B, Qwen2.5-0.5B, and Qwen2.5-1.5B for comparison. The Qwen series of small LLMs have been pre-trained on 18T tokens, and the training details have not been fully publicly released. They demonstrate strong performance in both general and domain-specific tasks.

    \item \textit{StableLM2-1.6B}~\citep{stablelm-2}:
StableLM2-1.6B is a small LLM proposed by StabilityAI. {It has been pre-trained on a mixture of open-source datasets, which utilizes several small LLMs to determine the training data proportion.} 

    \item \textit{SmolLM2-1.7B}~\citep{allal2024SmolLM2}:
SmolLM2-1.7B is developed by {HuggingFace TB Research based on its collected high-quality pre-training corpus}, which has been trained on 11T tokens, and maintains a good balance between speed and accuracy.

    \item \textit{Llama3.2-3B}~\citep{llama-3-card}:
Llama3.2-3B is developed by MetaAI, which is trained on up to 9T tokens. {It further distills the knowledge from LLaMA3.1-8B and 70B models by using their logits during the pretraining stage.}  
    \item \textit{Gemma2-2.6B}~\citep{gemma_2024}: Gemma2-2.6B is developed by Google, which is trained on 2T tokens, mainly including web documents, code, and mathematical text.
\end{itemize}



\subsubsection{Implementation Details}  


To comprehensively compare the performance of different LLMs, we employ diverse evaluation settings and design specific methods for guaranteeing the fairness and efficiency.

\begin{itemize}
    \item \textit{Zero-shot and few-shot settings:} Following existing work~\citep{qwen2.5}, 
For LAMBADA, HumanEval, MBPP, RACE, StoryCloze and RULER, we adopt the zero-shot setting.
For GSM8K and MATH, we adopt the 4-shot setting.  
For MMLU, CMMLU, WinoGrande and CEval, we adopt the 5-shot setting. 
For HellaSwag, we adopt the 10-shot setting. 
For ARC-E, ARC-C, we adopt the 25-shot setting. 

    \item \textit{Chain-of-Thought (CoT):} 
For GSM8K and MATH, we follow previous work~\citep{qwen2.5} that uses CoT prompting to facilitate the LLM to perform step-by-step reasoning. 
Considering the potential performance variance caused by CoT prompts, we utilize both the short ones provided by the original dataset and the long ones generated by \texttt{kimi-k0-math}. For each model, we evaluate the performance using both prompt types, and select the one yielding the higher score as the result. 

    \item \textit{Evaluation metrics:}
For QA tasks, we employ \texttt{maj@1} for GSM8K and MATH, \texttt{pass@1} for HumanEval and MBPP, and accuracy of the model response for remaining generation tasks. For multiple-choice questions, we primarily evaluate based on the accuracy of the generated answer, which is determined by selecting the choice with the lowest perplexity. However, for ARC-E and ARC-C, we utilize normalized accuracy~\citep{brown2020languagemodelsfewshotlearners}. 
To accurately measure the performance of MATH-500, we further use \texttt{gpt-4o-mini} to verify the correctness of the results generated by all models and conducted manual checks.

    \item \textit{Maximum length:} 
For GSM8K and MATH, since CoT prompting may result in longer outputs, we set the maximum generation length to 596 for short context (\ie 4K) models and 2,048 for long context models. For HumanEval and MBPP, we set the maximum generation length to 400.
For other generative tasks, we set it to 128 for efficiency.

    \item\textit{Evaluation framework:}
For the majority of tasks, we employ LLMBox~\citep{tang_llmbox_2024} to assess performance. Specifically, for generation tasks, we enable vLLM~\citep{kwon_efficient_2023}. However, to ensure reproducibility, we utilize EvalPlus~\citep{evalperf} for HumanEval and MBPP.

\end{itemize}

Despite our considerable efforts, fully reproducing the results of these baseline models as originally reported remains challenging, due to the lack of detailed evaluation setup information. 
For a fair comparison, we report the performance results of the baselines as provided in their official technical reports.

\subsection{Main Results}

\begin{table*}[t]
    \centering
    \small
    \caption{Performance on math, code, and long context benchmarks. Results marked with * are cited from their official paper or report. The best and second best results are \textbf{bold} and \underline{underlined}, respectively.}
    \setlength\tabcolsep{1.2mm}{
    \begin{tabular}{lrrr|ccccccc}
    \toprule
    \textbf{Models} & \tabincell{c}{\textbf{Model} \\ \textbf{Size} }& \tabincell{c}{\textbf{\# Train} \\ \textbf{Tokens}}& \tabincell{c}{\textbf{Context} \\ \textbf{Length} } & \tabincell{c}{\textbf{MATH}\\ \textbf{500}} & \tabincell{c}{\textbf{GSM}\\ \textbf{8K}} & \tabincell{c}{\textbf{Human}\\\textbf{Eval}} & \tabincell{c}{\textbf{MBPP}} &  \tabincell{c}{\textbf{RACE} \\ \textbf{Middle}}& \tabincell{c}{\textbf{RACE}\\ \textbf{High}}  & \tabincell{c}{\textbf{RULER}  } \\
    \midrule
    MiniCPM & 2.6B & 1.06T & 4K & 15.00\hspace*{1.5mm} & 53.83\hspace*{1.5mm} & 50.00$^*$ & 47.31\hspace*{1.5mm} & 56.61 & 44.27 & N/A \\
    Qwen-2 & 1.5B & 7T & 128K & 22.60\hspace*{1.5mm} & 46.90$^*$ & 34.80$^*$ & 46.90$^*$ & 55.77 & 43.69 & 60.16 \\
    Qwen2.5 & 0.5B & 18T & 128K & 23.60\hspace*{1.5mm} & 41.60$^*$ & 30.50$^*$ & 39.30$^*$ & 52.36 & 40.31 & 49.23 \\
    Qwen2.5 & 1.5B & 18T & 128K & \textbf{45.40}\hspace*{1.5mm} & \textbf{68.50}$^*$ & 37.20$^*$ & {60.20}$^*$ & \textbf{58.77} & 44.33 & \underline{68.26} \\
    Gemma2 & 2.6B & 2T & 8K & 18.30$^*$ & 30.30$^*$ & 19.50$^*$ & 42.10$^*$ & - & - & N/A \\
    StableLM2 & 1.7B & 2T & 4K & - & 20.62\hspace*{1.5mm} & 8.50 & 17.50\hspace*{1.5mm} & 56.33 & \textbf{45.06} & N/A \\
    SmolLM2 & 1.7B & 11T & 8K & 11.80\hspace*{1.5mm} & - & 23.35\hspace*{1.5mm} & 45.00\hspace*{1.5mm} & 55.77 & 43.06 & N/A \\
    Llama3.2 & 3.2B & 9T & 128K & 7.40 & - & 29.30\hspace*{1.5mm} & 49.70\hspace*{1.5mm} & 55.29 & 43.34 & \textbf{77.06} \\
    \midrule
    \multirow{2}{*}{YuLan-Mini} & 2.4B & 1.04T & 4K & 32.60\hspace*{1.5mm} & 66.65\hspace*{1.5mm} & \underline{61.60}\hspace*{1.5mm} & \textbf{66.70}\hspace*{1.5mm} & 55.71 & 43.58 & N/A \\
     & 2.4B & 1.08T & 28K & \underline{37.80}\hspace*{1.5mm} & \underline{68.46}\hspace*{1.5mm} & \textbf{64.00}\hspace*{1.5mm} & \underline{65.90}\hspace*{1.5mm} & \underline{57.18} & \underline{44.57} & 51.48 \\
    \bottomrule
    \end{tabular}}
    \label{tab:math_code_benchmark}
\end{table*}

\begin{table*}[t]
    \centering
    \small
    \caption{Performance on commonsense reasoning benchmarks. Results marked with * are cited from their official paper or report.}
    \setlength\tabcolsep{1.3mm}{
    \begin{tabular}{lcccccccccc}
    \toprule
    \textbf{Models} & \tabincell{c}{\textbf{LAMBADA}}  & \tabincell{c}{\textbf{MMLU}}  & \tabincell{c}{\textbf{CMMLU}}  & \tabincell{c}{\textbf{CEval}}  & \tabincell{c}{\textbf{Hella} \\ \textbf{Swag}} &  \tabincell{c}{\textbf{Wino} \\ \textbf{Grande}}  & \tabincell{c}{\textbf{Story} \\ \textbf{Cloze}} & \tabincell{c}{\textbf{ARC-e}}  & \tabincell{c}{\textbf{ARC-c}} \\
    \midrule
    MiniCPM-2.6B & 61.91 & 53.37\hspace*{1.5mm} & 48.97 & 48.24\hspace*{1.5mm} & 67.92\hspace*{1.5mm} & 65.74\hspace*{1.5mm} & 78.51 & 55.51 & 43.86\hspace*{1.5mm} \\
    
    Qwen2-1.5B & 64.68 & 55.90\hspace*{1.5mm} & \textbf{70.76} & \textbf{71.94}\hspace*{1.5mm} & 66.11\hspace*{1.5mm} & 66.14\hspace*{1.5mm} & 77.60 & {62.21} & 42.92\hspace*{1.5mm} \\
    
    Qwen2.5-0.5B & 52.00 & 47.50\hspace*{1.5mm} & 52.17 & 54.27\hspace*{1.5mm} & 50.54\hspace*{1.5mm} & 55.88\hspace*{1.5mm} & 71.67 & 56.10 & 39.51\hspace*{1.5mm} \\
    
    Qwen2.5-1.5B & 62.12 & \underline{60.71}\hspace*{1.5mm} & \underline{67.82} & \underline{69.05}\hspace*{1.5mm} & 67.18\hspace*{1.5mm} & 64.48\hspace*{1.5mm} & 76.80 & \textbf{71.51} & \underline{53.41}\hspace*{1.5mm} \\
    
    Gemma2-2.6B & - & 52.20$^*$ & - & {28.00}$^*$ & \underline{74.60}$^*$ & \textbf{71.50}$^*$ & - & - & \textbf{{55.70}}$^*$ \\
    
    StableLM2-1.7B & 66.15 & 40.37\hspace*{1.5mm} & 29.29 & 26.99\hspace*{1.5mm} & 69.79\hspace*{1.5mm} & 64.64\hspace*{1.5mm} & \underline{78.56} & 54.00 & 40.78\hspace*{1.5mm} \\
    
    SmolLM2-1.7B & \underline{67.42} & 51.91\hspace*{1.5mm} & 33.46 & 35.10\hspace*{1.5mm} & {72.96}\hspace*{1.5mm} & 67.40\hspace*{1.5mm} & \textbf{79.32} & 44.82 & 35.49\hspace*{1.5mm} \\
    
    Llama3.2-3B & \textbf{69.08} & \textbf{63.40}\hspace*{1.5mm} & 44.44 & 44.49\hspace*{1.5mm} & \textbf{75.62}\hspace*{1.5mm} & \underline{67.48}\hspace*{1.5mm} & 76.80 & \underline{70.12} & 48.81\hspace*{1.5mm} \\
    
    \midrule
    \multirow{2}{*}{YuLan-Mini} & 64.72 & 51.79\hspace*{1.5mm} & 48.35 & 51.47\hspace*{1.5mm} & {68.65}\hspace*{1.5mm} & 67.09\hspace*{1.5mm} & 76.37 & 69.87 & {50.51}\hspace*{1.5mm} \\
    
    & 65.67 & 49.10\hspace*{1.5mm} & 45.45 & 48.23\hspace*{1.5mm} & 67.22\hspace*{1.5mm} & 67.24\hspace*{1.5mm} & 75.89 & 67.47 & 49.32\hspace*{1.5mm} \\
    \bottomrule
    \end{tabular}}
    \label{tab:others_benchmark}
\end{table*}

The experimental results of different models on the specified  benchmarks are shown in Table~\ref{tab:math_code_benchmark} and Table~\ref{tab:others_benchmark}.
Furthermore, we have selected 4 benchmarks from each of the two tables to construct Figure~\ref{fig:performance_compare}.
Based on these results, we can identify the following key observations: 

\paragraph{Superior training efficacy} 
Overall, YuLan-Mini achieves highly competitive performance compared to leading small industry models, despite being trained on a significantly smaller corpus (1.08T tokens). 
To ensure successful pre-training with relatively limited data, we meticulously designed the data pipeline, effectively mitigated training instability, and implemented annealing training. 
Additionally, most of our training data comes from open-source and synthetic datasets, demonstrating that with careful data cleaning, selection, and scheduling, we can develop a robust base model even with limited resources in a university-level laboratory setting. 
This highlights the superior data efficiency of our pre-training approach. 

\paragraph{Excellence in mathematical and coding} On specific benchmarks for mathematical reasoning (MATH-500 and GSM8K) and coding generation (HumanEval and MBPP), YuLan-mini achieves leading average performance. This consistent superiority can be mainly attributed to the use of high-quality pre-training corpus and reasoning synthetic data (\eg formal mathematics reasoning problems and o1-like long thought data).
Our core idea is to extend the types of reasoning data and enhance the complex reasoning capacities of our base model, which leads to large improvements on mathematical benchmarks. 

\paragraph{Strong general capability} Beyond specialized tasks, YuLan-mini also demonstrates strong performance on various general benchmarks, spanning from language modeling and commonsense reasoning, highlighting the versatility of the model. It indicates that our pre-training approach  well balances the learning of diverse abilities, resulting in a robust general-purpose foundation model. This success can be attributed to our data mixture and curriculum-based adjustment strategies, which carefully balance the initial proportion of math, coding, and general knowledge related corpora, while gradually enhancing them during the annealing stage to better develop advanced capabilities. 

\paragraph{Moderate long context capability} 
Due to limited computing resources, YuLan-Mini has had limited exposure to long context training samples. As a result, its ability to model long contexts is not yet on par with state-of-the-art LLMs, as reflected in the results from the RULER benchmark. Additionally, due to a lack of GPU resources, our current approach can only extend the context window up to 28K.  This limitation could be addressed with additional resources to support further development.

\subsection{Evaluating Model Performance during Pre-Training}\label{sec:monitor}



\begin{figure}[t]
    \centering
    \begin{subfigure}[b]{0.49\textwidth}
        \centering
        \includegraphics[width=\textwidth]{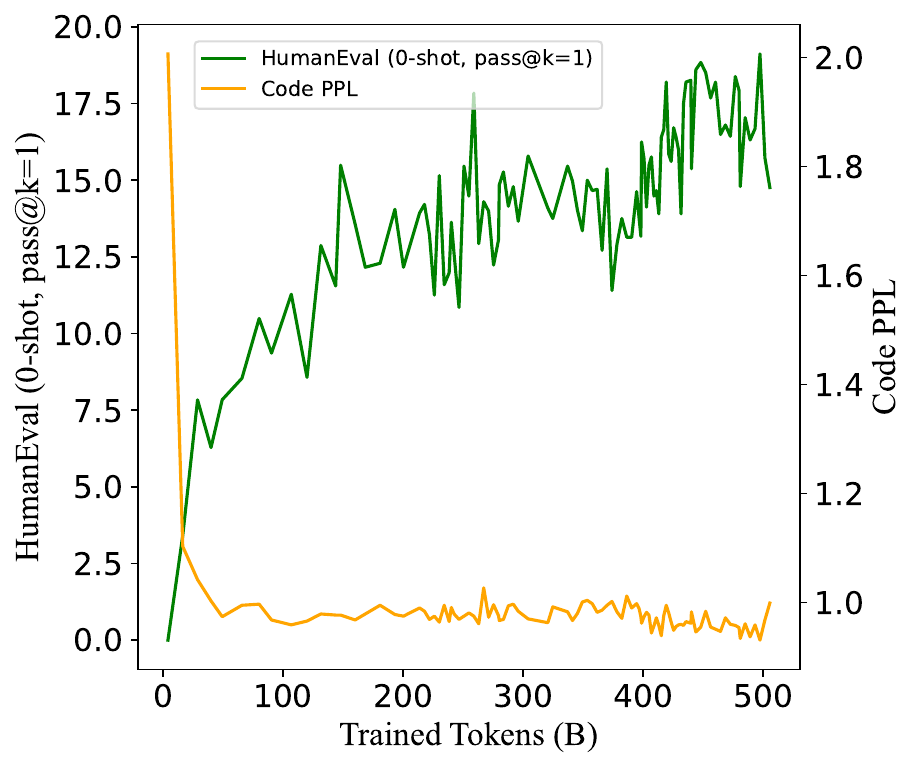}
        \caption{Performance curve on HumanEval.}
        \label{fig:monitor_code}
    \end{subfigure}
    \begin{subfigure}[b]{0.49\textwidth}
        \centering
        \includegraphics[width=\textwidth]{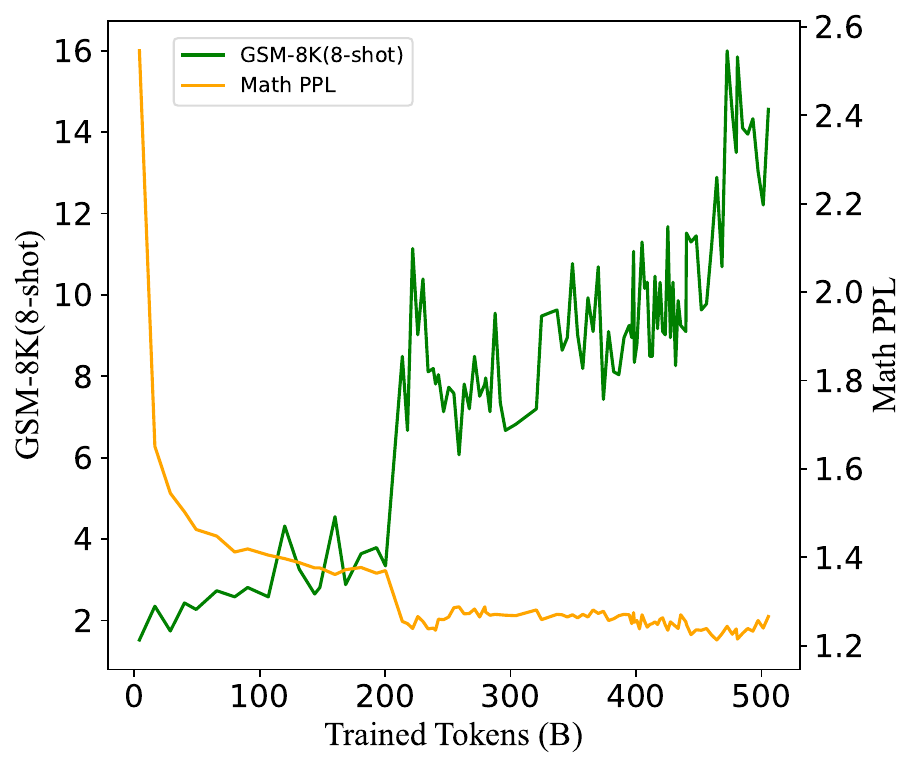}
        \caption{Performance curve on GSM8K.}
        \label{fig:monitor_math}
    \end{subfigure}
    \caption{Performance comparison using perplexity (PPL) and accuracy-based metrics to monitor the code generation and math reasoning abilities of YuLan-Mini.}
    \label{fig:monitor}
\end{figure}

During pre-training, it is crucial to continuously evaluate the model's performance to monitor for any unstable or abnormal training issues.
However, existing benchmarks rely on advanced abilities (\eg instruction following), which often develop with sufficient data training. Thus, the model's performance tends to remain at a low level on these benchmarks in the early stages, and directly evaluating the model's performance on specific validation sets would not provide an accurate assessment.

To address this, we have designed two monitoring strategies for different stages of training. In the early stages, we assess the model's performance primarily through perplexity measures on the constructed validation datasets and LAMBADA benchmark. In the later stages, we shift to using performance on selected benchmarks (\eg HumanEval and GSM8K) for more comprehensive evaluation. Next, we introduce how to construct the validation set for perplexity measurement at early stage of pre-training.

To comprehensively evaluate the key abilities of our model, we create four validation sets from the following aspects, namely English understanding, Chinese understanding, code generation, and math reasoning. The detailed data composition is as follows.

\begin{itemize}
    \item \textit{English understanding}: We randomly select 2,118 samples from FineWeb-Edu and compute the perplexity for ability evaluation. 

    \item \textit{Chinese understanding}: We randomly select 1,679 samples from {Chinese-FineWeb-Edu} for computing the perplexity. 

    \item \textit{Code generation}: We randomly select 2,067 samples from a widely-used code instruction datasets, {Python-Code-Instructions-18k-Alpaca} for perplexity evaluation.\footnote{https://huggingface.co/datasets/iamtarun/python\_code\_instructions\_18k\_alpaca} 

    \item \textit{Math reasoning}: We randomly sample 1,499 open-ended questions from MathInstruct~\citep{yue2023mammoth} for perplexity. 
\end{itemize}

Once the advanced capabilities are well-developed, we can directly monitor the model's performance by evaluating it on the selected benchmarks.

\section{Conclusion}
In this paper, we have introduced YuLan-Mini, a highly capable base model with 2.42B parameters. We provide all the essential technical details and resources necessary to reproduce our model, including improvements to the training method for enhanced stability and efficiency, as well as tokenized data sources organized with a specially developed training  curriculum. Extensive experiments have demonstrated the effectiveness of YuLan-Mini, showing that it achieves performance comparable to its industry counterparts of a similar parameter scale. Our primary contribution is enabling the reproduction of competitive language models in a highly data-efficient manner, making it feasible for university-level laboratories. 
Additionally, by aligning the training data with intermediate checkpoints, our approach facilitates in-depth research on LLMs, such as exploring how model capacities develop during pre-training.


 In future work, we plan to release an instruct version of YuLan-Mini. Furthermore, we aim to extend YuLan-Mini to other architectures and training methods and also explore its specialization in professional domains (\eg math and coding).

\section*{Acknowledgment} 
We sincerely thank Weizheng Lu and  Xu Han  for their assistance with this work.  
We encourage more researchers to collaborate in uncovering the pre-training secrets of LLMs. 








\bibliographystyle{plainnat}
\footnotesize
\bibliography{references.bib}
\normalsize

\balance
\newpage

\appendix

\section{Definition of Variables}
We list the detailed definition of all the used variables from Section~\ref{sec:training_stability} in Table~\ref{tab:stability_recipe_explain}.

\begin{table*}[h]
    \centering
    \small
    \caption{Definition of the variables for computing the hyperparameters.}
    \renewcommand{\arraystretch}{1.4}
    \begin{center}
    {
    \begin{tabular}{ll}
    \toprule
    \textbf{Variables}  &   \textbf{Meaning} \\  
    \midrule[0.5pt]
    $n_{\text{layers}} $& The num of model's layers \\
    $n_{\text{heads}} $& The num of model's attention heads \\
    $n_{\text{kv\_heads}} $& The num of model's kv-heads used in GQA \\
    
    $d_{\text{model}}$& Model dimension, \ie hidden size \\
    $d_{\text{head}} $& Dimension of attention head \\
    $d_{\text{ffn}} $&  The hidden size of feed-forward network \\
    $\sigma_{\text{base}}$& Initialization standard deviation for each matrix \\
    $\eta_{\text{base}}$& Learning rate, \ie max learning rate \\
    \midrule
    $d_{\text{model\_proxy}}$ & $d_{\text{model}}$ for proxy model, \ie the 0.05B model \\
    $m_{\text{width}}$& Width scaling factor in $\mu$P, \ie $d_{\text{model}} / d_{\text{model\_proxy}}$ \\
    \bottomrule
    \end{tabular}}
    \end{center}
    \label{tab:stability_recipe_explain}
\end{table*}

\section{Training Stability}\label{sec:appendix-training-stability}

\subsection{Experiment Setup}\label{sec:appendix-training-stability-experiment-setup}

We employ a relatively large learning rate with the intention of revealing the instability within the model. Unless otherwise specified, a trial learning rate of 0.01 is adopted.
We sample 20B tokens from our pre-training dataset, ensuring consistency with the source used in the final training.

The detailed architecture settings are listed in Table~\ref{tab:exploring-architecture}. The 0.2B proxy model is utilized for general exploratory experiments, whereas the 0.05B and 0.4B model are employed for $\mu$P experiments. The latter proxy model has the same number of layers as the final model.

\begin{table}[h]
    \centering
    \small
    \caption{Small proxy models used to explore the training dynamics.}
    \begin{center}
    \begin{tabular}{lcccccc}
    \toprule
    \textbf{Model} & {LR} & {$n_\text{layers}$} & {$d_\text{model}$} & {$d_\text{ffn}$} & {$n_\text{heads}$} & {$n_\text{kv\_heads}$} \\
    \midrule[0.5pt]
    YuLan-Mini & 0.01 & 56 & 1920 & 4800 & 30 & 6 \\
    \midrule
    Proxy model (0.05B) & 0.01 & 32 & 256 & 640 & 2 & 2 \\
    Proxy model (0.2B) & 0.01 & 30 & 576 & 1536 & 9 & 3 \\
    Proxy model (0.4B) & 0.01 & 56 & 576 & 1536 & 9 & 3 \\
    \bottomrule
    \end{tabular}
    \end{center}
    \label{tab:exploring-architecture}
\end{table}

\section{Prompts for Generating Synthetic Data}\label{sec:appendix-prompt}

\subsection{Math}

\begin{promptbox}[Prompt for Mathematical Documents Synthesis A]{lightblue}
\#\# Instruction\\
Please gain inspiration from the following content to create a lecture script for a college-level mathematics course.\\
\\
\#\# Content\\
\{Content Placeholder\}
\\
Begin without using titles or introductions.
\end{promptbox}

\begin{promptbox}[Prompt for Mathematical Documents Synthesis B]{lightblue}
Write an educational piece suited for college students related to the following text snippet:\\
\{Content Placeholder\}\\
\\
Do not just list concepts, but develop each one in detail before moving to the next, as we prioritize depth of understanding and comprehensive exploration of the subject matter over breadth. Focus on:\\
\\
- Rigor: Ensure in-depth coverage of the concepts/sections.\\
- Engagement: Write with an academic, professional and engaging tone that captivates interest.\\
- Application: Incorporate specific, practical examples, such as proofs in calculus or critical dates and figures in history.\\
\\
Do not include a title or an introduction, simply write the content without headlines and introductory phrases.
\end{promptbox}

\begin{promptbox}[Prompt for Mathematical Documents Synthesis C]{lightblue}
Write an educational piece suited for middle school students related to the following text snippet:\\
\{Content Placeholder\}\\
\\
Do not just list concepts, but develop each one in detail before moving to the next, as we prioritize depth of understanding and comprehensive exploration of the subject matter over breadth. Focus on:\\
\\
- Rigor: Ensure in-depth coverage of the concepts/sections.\\
- Engagement: Write with an academic, professional and engaging tone that captivates interest.\\
- Application: Incorporate specific, practical examples, such as proofs in calculus or critical dates and figures in history.\\
\\
Do not include a title or an introduction, simply write the content without headlines and introductory phrases.\\
\end{promptbox}

\begin{promptbox}[Prompt for Mathematical Documents Synthesis D]{lightblue}
\#\# Instruction\\
Please gain inspiration from the following content to draft a mathematics textbook chapter suitable for college students:\\
\\
\#\# Content\\
\{Content Placeholder\}\\
\\
Write in a clear, structured manner that is easy for students to follow and understand.
\end{promptbox}

\begin{promptbox}[Prompt for Mathematical Documents Synthesis E]{lightblue}
\#\# Instruction\\
Please gain inspiration from the following content to draft a mathematics textbook chapter suitable for middle school students:\\
\\
\#\# Content\\
\{Content Placeholder\}\\
\\
Write in a clear, structured manner that is easy for students to follow and understand.
\end{promptbox}

\begin{promptbox}[Prompt for Mathematical Documents Synthesis F]{lightblue}
\#\# Instruction\\
Please gain inspiration from the following content to design a problem set with solutions.\\
\\
\#\# Content\\
\{Content Placeholder\}\\
\\
\#\# Guidelines\\
- Formulate a series of problems that test understanding and application of the concepts.\\
- Provide detailed solutions for each problem, explaining the reasoning and calculations involved.\\
- Vary the difficulty of the problems to cater to students with different levels of proficiency.\\
\\
The problem set should challenge students to apply their knowledge in practical scenarios.
\end{promptbox}

\begin{promptbox}[Prompt for Mathematical Instruction Synthesis]{lightblue}
You are exceptionally skilled at crafting high-quality math problems and offering precise solutions.\\\\
Please gain inspiration from the following random math content to create a high-quality math problem and solve it step by step with clear logic. Present your output in two distinct sections:\\\\

[Problem Description] and [Solution]\\\\

Math content for inspiration:\\\\
\{Seed Content Placeholder\}\\

Guidelines for each section:\\\\

1. [Problem Description]: This should be **completely self-contained**, providing all the contextual information one needs to understand and solve the problem.\\
2. [Solution]: Offer a comprehensive, **correct** solution that accurately addresses the [Problem Description] you provided step by step with clear logic. Please ensure that the Solution only involves answering the Problem, **without addressing the requirements I provided.**
\end{promptbox}

\subsection{Code}

\begin{promptbox}[Prompt for Code Instruction Synthesis A]{lightblue}
You are a teaching assistant helping to create a Python programming task from a given code snippet. You must provide the best response to the Python programming task, including reasoning thought and reference solutions.\\

[Code Snippet]

\{Code Placeholder\}
\\
Your response must have these parts:\\

[Task]

\{Create an independent and detailed Python programming task\}\\

[Analysis]

\{Analyze the task and reason about the given task step by step\}\\

[Solution]

\{Write a high-quality reference solution in a self-contained script that solves the task\}
\end{promptbox}

\begin{promptbox}[Prompt for Code Instruction Synthesis B]{lightblue}
You are exceptionally skilled at crafting high-quality Python programming problems and offering precise solutions.
\\

Please gain inspiration from the following random code snippet to create a high-quality programming problem. Present your output in two distinct sections:
\\

[Problem Description] and [Solution]
\\

Code snippet for inspiration:
\\
\{Code Placeholder\}
\\\\
Guidelines for each section:
\\
1. [Problem Description]: This should be **completely self-contained**, providing all the contextual information one needs to understand and solve the problem. Assume common programming knowledge, but ensure that any specific context, variables, or code snippets pertinent to this problem are explicitly included.\\
2. [Solution]: Offer a comprehensive, **correct** solution that accurately addresses the [Problem Description] you provided. Please ensure that the Solution only involves answering the Problem, **without addressing the requirements I provided!**
\end{promptbox}

\subsection{Science}

\begin{promptbox}[Prompt for Scientific QA Synthesis]{lightblue}
\textbf{Instruction}

Please gain inspiration from the following \{Discipline Placeholder\} content to create a high-quality \{Discipline Placeholder\} problem and solution. Present your output in two distinct sections: [Problem] and [Solution].\\

\textbf{\{Discipline Placeholder\} Content}

\{Seed Snippet Placeholder\}\\

\textbf{Guidelines}

[Problem]: This should be **completely self-contained**, providing all the contextual information one needs to understand and solve the problem.\\

[Solution]: Present a comprehensive, step-by-step solution that solves the problem **correctly** and educates the student, around 250-350 words long. Clearly articulate the reasoning and methods used at each step, providing insight into the problem-solving process. Take care to format any equations properly using LaTeX or appropriate notation.\\
\end{promptbox}

\begin{promptbox}[Prompt for Topic Labeling]{lightblue}
I am categorizing a series of articles according to the following 11 topics. Next, I will give you an article, please select only one topic that the article is the most related to:\\

[Topics]: \{Topic List Placeholder\}\\

[Article]: \{Web Page Content Placeholder\}\\

Please only return the most related topic:
\end{promptbox}

\subsection{Data Selection}

\begin{promptbox}[Prompt for Instruction Tag Labeling]{lightblue}
Please identify the relevant tags representing the user's intentions in the following Problem and Solution. Focus on the reasoning behind the solution. Please ONLY respond with tags in a Python list.\\

Problem:\\
\{Question Placeholder\}
\\
Solution:\\
\{Answer Placeholder\}\
\end{promptbox}

\section{Open-Source Datasets Used during Pre-training}\label{sec:data-list}

{
\small
\footnotesize
\captionof{table}{Comprehensive list of all open-source datasets used. For datasets that are only available via links, we also offer additional guidance on our project website~\url{https://github.com/RUC-GSAI/YuLan-Mini}.}

\begin{longtable}{p{2cm}p{8.2cm}}
\toprule
\textbf{Domain} & \textbf{Dataset}\\
\endfirsthead
\midrule
General & \tabincell{l}{
\texttt{chinese-fineweb-edu}~\citep{OpencsgChinesefinewebeduDatasets} \\
\texttt{llm360-txt360}~\citep{txt360data2024} \\
\texttt{wanjuan}~\citep{he_wanjuan_2023} \\
\texttt{dclm}~\citep{dclm-report} \\
\texttt{fineweb-edu}~\citep{penedo_fineweb_2024} \\
\texttt{dolma}~\citep{soldaini_dolma_2024} \\
\texttt{tulu3}~\citep{lambert2024tulu3} \\
\texttt{magpie-reasoning-150k}~\citep{xu_magpie_2024}}  \\

\addlinespace
\hline
\addlinespace

Code & \tabincell{l}{
\texttt{the-stack-v2}~\citep{lozhkov_starcoder_2024}\\
\texttt{starcoderdata}~\citep{li_starcoder_2023}\\ 
\texttt{opencoder-llm}~\citep{huang_opencoder_2024}\\
\texttt{longwanjuan-github}~\citep{lv_longwanjuan_2024}\\
\texttt{mathcodeinstruct}~\citep{wang_mathcoder_2023}\\
\texttt{codefeedback-filtered-instruction}~\citep{zheng2024opencodeinterpreterintegratingcodegeneration}\\
\texttt{xcoder-80k}~\citep{wang_how_2024}}\\
\addlinespace
\hline
\addlinespace

Math & \tabincell{l}{
\texttt{proof-pile-2}~\citep{azerbayev_llemma_2024}\\
\texttt{automathtext}~\citep{zhang_autonomous_2024}\\
\texttt{open-web-math-pro}~\citep{zhou2024programmingexampleliftingpretraining}\\
\texttt{deepmind-math}~\citep{deepmind-math}\\
\texttt{orca-math}~\citep{orca-math}\\
\texttt{metamathqa}~\citep{metamathqa}\\
\texttt{numina}~\citep{beeching_numinamath_2024}\\
\texttt{silmorca}~\citep{slimorca}\\
\texttt{scalequest-math}~\citep{ding_unleashing_2024}\\
\texttt{infimm-webmath-40b}~\citep{han_infimm-webmath-40b_2024}\\
\texttt{lean-star}~\citep{lin2024leanstarlearninginterleavethinking}\\
\texttt{lean-github}~\citep{wu_lean-github_2024}\\
\texttt{lean-workbook}~\citep{ying_lean_2024}\\
\texttt{lean-deepseek-v1}~\citep{xin_deepseek-prover_2024}\\
\texttt{ape210k}~\citep{kadlcik_calc-x_2023}\\
\texttt{mathinstruct}~\citep{yue2023mammoth}\\
\texttt{openmathinstruct-1}~\citep{toshniwal_openmathinstruct-1_2024}\\
\texttt{mathscaleqa-2m}~\citep{tang_mathscale_2024}\\
\texttt{orca-agentinstruct}~\citep{mitra_agentinstruct_2024}\\
\texttt{fol-nli}~\citep{sileo_scaling_2024}\\
\texttt{gretel-math-gsm8k-v1}~\citep{gretelai_gsm8k_synthetic}} \\
\addlinespace
\hline
\end{longtable}\label{tab:dataset_source}

}

\section{Detailed Data Composition by Training Phases}\label{data-all-data}

The specific composition of data for each course stage is shown in Table~\ref{tab:comp_by_phase}, where \texttt{black} represents English web pages and general content, \textcolor{darkgreen}{\texttt{green}} represents Chinese, \textcolor{navy}{\texttt{blue}} represents code, and \textcolor{darkred}{\texttt{red}} represents mathematics. 
The first 10 billion tokens of Phase 1 are used during the warm-up stage. The next 30 billion tokens of Phase 1, along with Phases 2 through 25, are employed in the stable training stage. Phases 26 and 27 constitute the annealing stage.
{
\small
\captionof{table}{Detailed data composition by training curriculum phases.}
\begin{longtable}{p{1cm}p{12cm}}
\toprule
\textbf{Phase}  & \textbf{Data composition by phase (in billions of tokens)}\\
\endfirsthead
\midrule
 \tabincell{l}{1}& \newtt{\textcolor{black}{{dclm}~(1.80)}, \textcolor{black}{{fineweb-edu}~(16.20)}, \textcolor{black}{{english-books}~(1.60)}, \textcolor{black}{{pes2o}~(0.80)}, \textcolor{black}{{arxiv}~(1.20)}, \textcolor{black}{{wikipedia}~(0.40)}, \textcolor{black}{{dolma}~(1.24)}, \textcolor{black}{{cicg-news}~(0.76)}, \textcolor{darkgreen}{{cn-baike}~(0.39)}, \textcolor{darkgreen}{{mnbvc-news}~(0.08)}, \textcolor{darkgreen}{{cn-book}~(0.24)}, \textcolor{darkgreen}{{cn-legal-case-law}~(0.36)}, \textcolor{darkgreen}{{zhihu-qa}~(0.12)}, \textcolor{navy}{{the-stack-v2}~(4.91)}, \textcolor{navy}{{starcoder}~(2.92)}, \textcolor{navy}{{smollm-python}~(0.20)}, \textcolor{darkred}{{proof-pile-2}~(1.52)}, \textcolor{darkred}{{automathtext}~(1.12)}, \textcolor{darkred}{{open-web-math-pro}~(0.20)}, \textcolor{darkred}{{cosmopedia}~(1.01)}, \textcolor{darkred}{{mathtext}~(0.12)}}\\
\addlinespace
\hline
\addlinespace
2 & \newtt{\textcolor{black}{{dclm}~(1.80)}, \textcolor{black}{{fineweb-edu}~(16.20)}, \textcolor{black}{{english-books}~(1.60)}, \textcolor{black}{{pes2o}~(0.80)}, \textcolor{black}{{arxiv}~(1.20)}, \textcolor{black}{{wikipedia}~(0.40)}, \textcolor{black}{{dolma}~(1.24)}, \textcolor{black}{{cicg-news}~(0.76)}, \textcolor{darkgreen}{{cn-baike}~(0.39)}, \textcolor{darkgreen}{{mnbvc-news}~(0.08)}, \textcolor{darkgreen}{{cn-book}~(0.24)}, \textcolor{darkgreen}{{cn-legal-case-law}~(0.36)}, \textcolor{darkgreen}{{zhihu-qa}~(0.12)}, \textcolor{navy}{{the-stack-v2}~(4.90)}, \textcolor{navy}{{starcoder}~(2.92)}, \textcolor{navy}{{smollm-python}~(0.20)}, \textcolor{darkred}{{proof-pile-2}~(1.52)}, \textcolor{darkred}{{automathtext}~(1.12)}, \textcolor{darkred}{{open-web-math-pro}~(0.20)}, \textcolor{darkred}{{cosmopedia}~(1.02)}, \textcolor{darkred}{{mathtext}~(0.12)}}\\
\addlinespace
\hline
\addlinespace
3 & \newtt{\textcolor{black}{{dclm}~(1.80)}, \textcolor{black}{{fineweb-edu}~(16.20)}, \textcolor{black}{{english-books}~(1.60)}, \textcolor{black}{{pes2o}~(0.80)}, \textcolor{black}{{arxiv}~(1.20)}, \textcolor{black}{{wikipedia}~(0.40)}, \textcolor{black}{{dolma}~(1.24)}, \textcolor{black}{{cicg-news}~(0.76)}, \textcolor{darkgreen}{{cn-baike}~(0.39)}, \textcolor{darkgreen}{{mnbvc-news}~(0.08)}, \textcolor{darkgreen}{{cn-book}~(0.24)}, \textcolor{darkgreen}{{cn-legal-case-law}~(0.36)}, \textcolor{darkgreen}{{zhihu-qa}~(0.12)}, \textcolor{navy}{{the-stack-v2}~(4.86)}, \textcolor{navy}{{starcoder}~(2.92)}, \textcolor{navy}{{smollm-python}~(0.20)}, \textcolor{darkred}{{proof-pile-2}~(1.52)}, \textcolor{darkred}{{automathtext}~(1.12)}, \textcolor{darkred}{{open-web-math-pro}~(0.20)}, \textcolor{darkred}{{cosmopedia}~(1.02)}, \textcolor{darkred}{{mathtext}~(0.12)}}\\
\addlinespace
\hline
\addlinespace
4 & \newtt{\textcolor{black}{{dclm}~(1.80)}, \textcolor{black}{{fineweb-edu}~(16.20)}, \textcolor{black}{{english-books}~(1.60)}, \textcolor{black}{{pes2o}~(0.80)}, \textcolor{black}{{arxiv}~(1.20)}, \textcolor{black}{{wikipedia}~(0.40)}, \textcolor{black}{{dolma}~(1.24)}, \textcolor{black}{{cicg-news}~(0.76)}, \textcolor{darkgreen}{{cn-baike}~(0.39)}, \textcolor{darkgreen}{{mnbvc-news}~(0.08)}, \textcolor{darkgreen}{{cn-book}~(0.24)}, \textcolor{darkgreen}{{cn-legal-case-law}~(0.36)}, \textcolor{darkgreen}{{zhihu-qa}~(0.12)}, \textcolor{navy}{{the-stack-v2}~(4.14)}, \textcolor{navy}{{starcoder}~(2.92)}, \textcolor{navy}{{smollm-python}~(0.20)}, \textcolor{darkred}{{proof-pile-2}~(1.52)}, \textcolor{darkred}{{automathtext}~(1.12)}, \textcolor{darkred}{{open-web-math-pro}~(0.24)}, \textcolor{darkred}{{cosmopedia}~(0.98)}, \textcolor{darkred}{{mathtext}~(0.12)}}\\
\addlinespace
\hline
\addlinespace
5 & \newtt{\textcolor{black}{{dclm}~(1.80)}, \textcolor{black}{{fineweb-edu}~(16.20)}, \textcolor{black}{{english-books}~(1.60)}, \textcolor{black}{{pes2o}~(0.80)}, \textcolor{black}{{arxiv}~(1.20)}, \textcolor{black}{{wikipedia}~(0.40)}, \textcolor{black}{{dolma}~(1.24)}, \textcolor{black}{{cicg-news}~(0.76)}, \textcolor{darkgreen}{{cn-baike}~(0.39)}, \textcolor{darkgreen}{{mnbvc-news}~(0.08)}, \textcolor{darkgreen}{{cn-book}~(0.24)}, \textcolor{darkgreen}{{cn-legal-case-law}~(0.36)}, \textcolor{darkgreen}{{zhihu-qa}~(0.12)}, \textcolor{navy}{{the-stack-v2}~(4.90)}, \textcolor{navy}{{starcoder}~(2.92)}, \textcolor{navy}{{smollm-python}~(0.20)}, \textcolor{darkred}{{proof-pile-2}~(1.54)}, \textcolor{darkred}{{automathtext}~(1.14)}, \textcolor{darkred}{{open-web-math-pro}~(0.24)}, \textcolor{darkred}{{cosmopedia}~(0.94)}, \textcolor{darkred}{{mathtext}~(0.12)}}\\
\addlinespace
\hline
\addlinespace
6 & \newtt{\textcolor{black}{{dclm}~(1.80)}, \textcolor{black}{{fineweb-edu}~(16.20)}, \textcolor{black}{{english-books}~(1.60)}, \textcolor{black}{{pes2o}~(0.80)}, \textcolor{black}{{arxiv}~(1.20)}, \textcolor{black}{{wikipedia}~(0.40)}, \textcolor{black}{{dolma}~(1.24)}, \textcolor{black}{{cicg-news}~(0.76)}, \textcolor{darkgreen}{{cn-baike}~(0.39)}, \textcolor{darkgreen}{{mnbvc-news}~(0.08)}, \textcolor{darkgreen}{{cn-book}~(0.24)}, \textcolor{darkgreen}{{cn-legal-case-law}~(0.36)}, \textcolor{darkgreen}{{zhihu-qa}~(0.12)}, \textcolor{navy}{{the-stack-v2}~(4.90)}, \textcolor{navy}{{starcoder}~(2.92)}, \textcolor{navy}{{smollm-python}~(0.20)}, \textcolor{darkred}{{proof-pile-2}~(1.54)}, \textcolor{darkred}{{automathtext}~(1.16)}, \textcolor{darkred}{{open-web-math-pro}~(0.26)}, \textcolor{darkred}{{cosmopedia}~(0.82)}, \textcolor{darkred}{{mathtext}~(0.12)}, \textcolor{darkred}{{metamathqa}~(0.02)}, \textcolor{darkred}{{orca-math}~(0.02)}, \textcolor{darkred}{{yulan-mini-syn-math-inst}~(0.04)}}\\
\addlinespace
\hline
\addlinespace
7 & \newtt{\textcolor{black}{{dclm}~(1.80)}, \textcolor{black}{{fineweb-edu}~(16.20)}, \textcolor{black}{{english-books}~(1.60)}, \textcolor{black}{{pes2o}~(0.80)}, \textcolor{black}{{arxiv}~(1.20)}, \textcolor{black}{{wikipedia}~(0.40)}, \textcolor{black}{{dolma}~(1.24)}, \textcolor{black}{{cicg-news}~(0.76)}, \textcolor{darkgreen}{{cn-baike}~(0.39)}, \textcolor{darkgreen}{{mnbvc-news}~(0.08)}, \textcolor{darkgreen}{{cn-book}~(0.24)}, \textcolor{darkgreen}{{cn-legal-case-law}~(0.36)}, \textcolor{darkgreen}{{zhihu-qa}~(0.12)}, \textcolor{navy}{{the-stack-v2}~(4.90)}, \textcolor{navy}{{starcoder}~(2.92)}, \textcolor{navy}{{smollm-python}~(0.20)}, \textcolor{darkred}{{proof-pile-2}~(1.60)}, \textcolor{darkred}{{automathtext}~(1.17)}, \textcolor{darkred}{{open-web-math-pro}~(0.28)}, \textcolor{darkred}{{cosmopedia}~(0.77)}, \textcolor{darkred}{{mathtext}~(0.12)}, \textcolor{darkred}{{metamathqa}~(0.01)}, \textcolor{darkred}{{orca-math}~(0.01)}, \textcolor{darkred}{{yulan-mini-syn-math-inst}~(0.02)}}\\
\addlinespace
\hline
\addlinespace
8 & \newtt{\textcolor{black}{{dclm}~(1.80)}, \textcolor{black}{{fineweb-edu}~(16.20)}, \textcolor{black}{{english-books}~(1.60)}, \textcolor{black}{{pes2o}~(0.80)}, \textcolor{black}{{arxiv}~(1.20)}, \textcolor{black}{{wikipedia}~(0.40)}, \textcolor{black}{{dolma}~(1.24)}, \textcolor{black}{{cicg-news}~(0.76)}, \textcolor{darkgreen}{{cn-baike}~(0.39)}, \textcolor{darkgreen}{{mnbvc-news}~(0.08)}, \textcolor{darkgreen}{{cn-book}~(0.24)}, \textcolor{darkgreen}{{cn-legal-case-law}~(0.36)}, \textcolor{darkgreen}{{zhihu-qa}~(0.12)}, \textcolor{navy}{{the-stack-v2}~(4.90)}, \textcolor{navy}{{starcoder}~(2.92)}, \textcolor{navy}{{smollm-python}~(0.20)}, \textcolor{darkred}{{proof-pile-2}~(1.64)}, \textcolor{darkred}{{automathtext}~(1.17)}, \textcolor{darkred}{{open-web-math-pro}~(0.32)}, \textcolor{darkred}{{cosmopedia}~(0.53)}, \textcolor{darkred}{{fineweb-math}~(0.16)}, \textcolor{darkred}{{mathtext}~(0.12)}, \textcolor{darkred}{{metamathqa}~(0.01)}, \textcolor{darkred}{{orca-math}~(0.01)}, \textcolor{darkred}{{yulan-mini-syn-math-inst}~(0.02)}}\\
\addlinespace
\hline
\addlinespace
9 & \newtt{\textcolor{black}{{dclm}~(1.80)}, \textcolor{black}{{fineweb-edu}~(16.20)}, \textcolor{black}{{english-books}~(1.20)}, \textcolor{black}{{pes2o}~(0.80)}, \textcolor{black}{{arxiv}~(1.20)}, \textcolor{black}{{wikipedia}~(0.40)}, \textcolor{black}{{dolma}~(1.24)}, \textcolor{black}{{cosmopedia-v2}~(0.40)}, \textcolor{black}{{cicg-news}~(0.76)}, \textcolor{darkgreen}{{cn-baike}~(0.39)}, \textcolor{darkgreen}{{mnbvc-news}~(0.08)}, \textcolor{darkgreen}{{cn-book}~(0.24)}, \textcolor{darkgreen}{{cn-legal-case-law}~(0.36)}, \textcolor{darkgreen}{{zhihu-qa}~(0.12)}, \textcolor{navy}{{the-stack-v2}~(4.86)}, \textcolor{navy}{{starcoder}~(2.92)}, \textcolor{navy}{{smollm-python}~(0.20)}, \textcolor{darkred}{{proof-pile-2}~(1.64)}, \textcolor{darkred}{{automathtext}~(1.17)}, \textcolor{darkred}{{open-web-math-pro}~(0.32)}, \textcolor{darkred}{{cosmopedia}~(0.33)}, \textcolor{darkred}{{fineweb-math}~(0.16)}, \textcolor{darkred}{{mathtext}~(0.12)}, \textcolor{darkred}{{metamathqa}~(0.01)}, \textcolor{darkred}{{orca-math}~(0.01)}, \textcolor{darkred}{{yulan-mini-syn-math-inst}~(0.02)}, \textcolor{darkred}{{yulan-mini-syn-math-doc}~(0.22)}}\\
\addlinespace
\hline
\addlinespace
10 & \newtt{\textcolor{black}{{dclm}~(1.80)}, \textcolor{black}{{fineweb-edu}~(16.20)}, \textcolor{black}{{english-books}~(1.00)}, \textcolor{black}{{pes2o}~(0.80)}, \textcolor{black}{{arxiv}~(1.20)}, \textcolor{black}{{wikipedia}~(0.40)}, \textcolor{black}{{dolma}~(1.24)}, \textcolor{black}{{cosmopedia-v2}~(0.60)}, \textcolor{black}{{cicg-news}~(0.76)}, \textcolor{darkgreen}{{cn-baike}~(0.39)}, \textcolor{darkgreen}{{mnbvc-news}~(0.08)}, \textcolor{darkgreen}{{cn-book}~(0.24)}, \textcolor{darkgreen}{{cn-legal-case-law}~(0.36)}, \textcolor{darkgreen}{{zhihu-qa}~(0.12)}, \textcolor{navy}{{the-stack-v2}~(4.85)}, \textcolor{navy}{{starcoder}~(2.92)}, \textcolor{navy}{{smollm-python}~(0.20)}, \textcolor{navy}{{yulan-mini-syn-code-inst}~(0.03)}, \textcolor{darkred}{{proof-pile-2}~(1.64)}, \textcolor{darkred}{{automathtext}~(1.17)}, \textcolor{darkred}{{open-web-math-pro}~(0.32)}, \textcolor{darkred}{{cosmopedia}~(0.29)}, \textcolor{darkred}{{fineweb-math}~(0.20)}, \textcolor{darkred}{{mathtext}~(0.12)}, \textcolor{darkred}{{metamathqa}~(0.01)}, \textcolor{darkred}{{orca-math}~(0.01)}, \textcolor{darkred}{{yulan-mini-syn-math-inst}~(0.02)}, \textcolor{darkred}{{yulan-mini-syn-math-doc}~(0.22)}}\\
\addlinespace
\hline
\addlinespace
11 & \newtt{\textcolor{black}{{dclm}~(1.80)}, \textcolor{black}{{fineweb-edu}~(16.20)}, \textcolor{black}{{english-books}~(0.82)}, \textcolor{black}{{pes2o}~(0.80)}, \textcolor{black}{{arxiv}~(1.20)}, \textcolor{black}{{wikipedia}~(0.40)}, \textcolor{black}{{dolma}~(1.24)}, \textcolor{black}{{cosmopedia-v2}~(0.78)}, \textcolor{black}{{cicg-news}~(0.76)}, \textcolor{darkgreen}{{cn-baike}~(0.39)}, \textcolor{darkgreen}{{mnbvc-news}~(0.08)}, \textcolor{darkgreen}{{cn-book}~(0.24)}, \textcolor{darkgreen}{{cn-legal-case-law}~(0.36)}, \textcolor{darkgreen}{{zhihu-qa}~(0.12)}, \textcolor{navy}{{the-stack-v2}~(4.56)}, \textcolor{navy}{{starcoder}~(2.92)}, \textcolor{navy}{{smollm-python}~(0.20)}, \textcolor{navy}{{mnbvc-code}~(0.04)}, \textcolor{navy}{{yulan-mini-syn-code-inst}~(0.28)}, \textcolor{darkred}{{proof-pile-2}~(1.64)}, \textcolor{darkred}{{automathtext}~(0.93)}, \textcolor{darkred}{{open-web-math-pro}~(0.41)}, \textcolor{darkred}{{cosmopedia}~(0.16)}, \textcolor{darkred}{{fineweb-math}~(0.33)}, \textcolor{darkred}{{dclm-math}~(0.12)}, \textcolor{darkred}{{mathtext}~(0.12)}, \textcolor{darkred}{{metamathqa}~(0.01)}, \textcolor{darkred}{{orca-math}~(0.01)}, \textcolor{darkred}{{yulan-mini-syn-math-inst}~(0.02)}, \textcolor{darkred}{{yulan-mini-syn-math-doc}~(0.25)}}\\
\addlinespace
\hline
\addlinespace
12 & \newtt{\textcolor{black}{{dclm}~(1.80)}, \textcolor{black}{{fineweb-edu}~(16.20)}, \textcolor{black}{{english-books}~(0.82)}, \textcolor{black}{{pes2o}~(0.80)}, \textcolor{black}{{arxiv}~(1.20)}, \textcolor{black}{{wikipedia}~(0.40)}, \textcolor{black}{{dolma}~(1.24)}, \textcolor{black}{{cosmopedia-v2}~(0.78)}, \textcolor{black}{{cicg-news}~(0.76)}, \textcolor{darkgreen}{{cn-baike}~(0.39)}, \textcolor{darkgreen}{{mnbvc-news}~(0.08)}, \textcolor{darkgreen}{{cn-book}~(0.24)}, \textcolor{darkgreen}{{cn-legal-case-law}~(0.36)}, \textcolor{darkgreen}{{zhihu-qa}~(0.12)}, \textcolor{navy}{{the-stack-v2}~(4.44)}, \textcolor{navy}{{starcoder}~(2.92)}, \textcolor{navy}{{smollm-python}~(0.20)}, \textcolor{navy}{{mnbvc-code}~(0.16)}, \textcolor{navy}{{yulan-mini-syn-code-inst}~(0.28)}, \textcolor{darkred}{{proof-pile-2}~(1.64)}, \textcolor{darkred}{{automathtext}~(0.63)}, \textcolor{darkred}{{open-web-math-pro}~(0.41)}, \textcolor{darkred}{{cosmopedia}~(0.03)}, \textcolor{darkred}{{fineweb-math}~(0.50)}, \textcolor{darkred}{{dclm-math}~(0.11)}, \textcolor{darkred}{{mathtext}~(0.12)}, \textcolor{darkred}{{basic-math-10m}~(0.04)}, \textcolor{darkred}{{metamathqa}~(0.01)}, \textcolor{darkred}{{orca-math}~(0.01)}, \textcolor{darkred}{{yulan-mini-syn-math-inst}~(0.06)}, \textcolor{darkred}{{yulan-mini-syn-math-doc}~(0.44)}}\\
\addlinespace
\hline
\addlinespace
13 & \newtt{\textcolor{black}{{dclm}~(1.80)}, \textcolor{black}{{fineweb-edu}~(16.20)}, \textcolor{black}{{english-books}~(0.82)}, \textcolor{black}{{pes2o}~(0.80)}, \textcolor{black}{{arxiv}~(1.20)}, \textcolor{black}{{wikipedia}~(0.40)}, \textcolor{black}{{dolma}~(1.24)}, \textcolor{black}{{cosmopedia-v2}~(0.78)}, \textcolor{black}{{cicg-news}~(0.76)}, \textcolor{darkgreen}{{cn-baike}~(0.39)}, \textcolor{darkgreen}{{mnbvc-news}~(0.08)}, \textcolor{darkgreen}{{cn-book}~(0.24)}, \textcolor{darkgreen}{{cn-legal-case-law}~(0.36)}, \textcolor{darkgreen}{{zhihu-qa}~(0.12)}, \textcolor{navy}{{the-stack-v2}~(4.44)}, \textcolor{navy}{{starcoder}~(2.92)}, \textcolor{navy}{{smollm-python}~(0.20)}, \textcolor{navy}{{mnbvc-code}~(0.16)}, \textcolor{navy}{{yulan-mini-syn-code-inst}~(0.28)}, \textcolor{darkred}{{proof-pile-2}~(1.64)}, \textcolor{darkred}{{automathtext}~(0.58)}, \textcolor{darkred}{{open-web-math-pro}~(0.41)}, \textcolor{darkred}{{fineweb-math}~(0.51)}, \textcolor{darkred}{{dclm-math}~(0.15)}, \textcolor{darkred}{{mathtext}~(0.12)}, \textcolor{darkred}{{basic-math-10m}~(0.04)}, \textcolor{darkred}{{metamathqa}~(0.01)}, \textcolor{darkred}{{yulan-mini-syn-math-inst}~(0.08)}, \textcolor{darkred}{{yulan-mini-syn-math-doc}~(0.44)}}\\
\addlinespace
\hline
\addlinespace
14 & \newtt{\textcolor{black}{{dclm}~(1.80)}, \textcolor{black}{{fineweb-edu}~(16.20)}, \textcolor{black}{{english-books}~(0.82)}, \textcolor{black}{{pes2o}~(0.80)}, \textcolor{black}{{arxiv}~(1.20)}, \textcolor{black}{{wikipedia}~(0.40)}, \textcolor{black}{{dolma}~(1.24)}, \textcolor{black}{{cosmopedia-v2}~(0.78)}, \textcolor{black}{{cicg-news}~(0.76)}, \textcolor{darkgreen}{{cn-baike}~(0.33)}, \textcolor{darkgreen}{{mnbvc-news}~(0.08)}, \textcolor{darkgreen}{{cn-book}~(0.24)}, \textcolor{darkgreen}{{cn-legal-case-law}~(0.36)}, \textcolor{darkgreen}{{zhihu-qa}~(0.12)}, \textcolor{navy}{{the-stack-v2}~(4.44)}, \textcolor{navy}{{starcoder}~(2.92)}, \textcolor{navy}{{smollm-python}~(0.20)}, \textcolor{navy}{{mnbvc-code}~(0.16)}, \textcolor{navy}{{yulan-mini-syn-code-inst}~(0.28)}, \textcolor{darkred}{{proof-pile-2}~(1.64)}, \textcolor{darkred}{{automathtext}~(0.57)}, \textcolor{darkred}{{open-web-math-pro}~(0.41)}, \textcolor{darkred}{{cosmopedia}~(0.02)}, \textcolor{darkred}{{fineweb-math}~(0.51)}, \textcolor{darkred}{{dclm-math}~(0.15)}, \textcolor{darkred}{{mathtext}~(0.12)}, \textcolor{darkred}{{basic-math-10m}~(0.04)}, \textcolor{darkred}{{metamathqa}~(0.01)}, \textcolor{darkred}{{yulan-mini-syn-math-inst}~(0.09)}, \textcolor{darkred}{{yulan-mini-syn-math-doc}~(0.44)}}\\
\addlinespace
\hline
\addlinespace
15 & \newtt{\textcolor{black}{{dclm}~(1.80)}, \textcolor{black}{{fineweb-edu}~(16.20)}, \textcolor{black}{{english-books}~(0.82)}, \textcolor{black}{{pes2o}~(0.80)}, \textcolor{black}{{arxiv}~(1.20)}, \textcolor{black}{{wikipedia}~(0.40)}, \textcolor{black}{{dolma}~(1.24)}, \textcolor{black}{{cosmopedia-v2}~(0.78)}, \textcolor{black}{{cicg-news}~(0.76)}, \textcolor{darkgreen}{{cn-baike}~(0.27)}, \textcolor{darkgreen}{{mnbvc-news}~(0.08)}, \textcolor{darkgreen}{{cn-book}~(0.24)}, \textcolor{darkgreen}{{cn-legal-case-law}~(0.36)}, \textcolor{darkgreen}{{zhihu-qa}~(0.12)}, \textcolor{navy}{{the-stack-v2}~(4.37)}, \textcolor{navy}{{starcoder}~(2.32)}, \textcolor{navy}{{smollm-python}~(0.20)}, \textcolor{navy}{{mnbvc-code}~(0.83)}, \textcolor{navy}{{yulan-mini-syn-code-inst}~(0.28)}, \textcolor{darkred}{{proof-pile-2}~(1.34)}, \textcolor{darkred}{{automathtext}~(0.65)}, \textcolor{darkred}{{open-web-math-pro}~(0.41)}, \textcolor{darkred}{{cosmopedia}~(0.02)}, \textcolor{darkred}{{fineweb-math}~(0.42)}, \textcolor{darkred}{{dclm-math}~(0.40)}, \textcolor{darkred}{{mathtext}~(0.12)}, \textcolor{darkred}{{basic-math-10m}~(0.04)}, \textcolor{darkred}{{metamathqa}~(0.01)}, \textcolor{darkred}{{yulan-mini-syn-math-inst}~(0.13)}, \textcolor{darkred}{{yulan-mini-syn-math-doc}~(0.46)}}\\
\addlinespace
\hline
\addlinespace
16 & \newtt{\textcolor{black}{{dclm}~(1.80)}, \textcolor{black}{{fineweb-edu}~(16.20)}, \textcolor{black}{{english-books}~(0.82)}, \textcolor{black}{{pes2o}~(0.80)}, \textcolor{black}{{arxiv}~(1.20)}, \textcolor{black}{{wikipedia}~(0.40)}, \textcolor{black}{{dolma}~(1.24)}, \textcolor{black}{{cosmopedia-v2}~(0.78)}, \textcolor{black}{{cicg-news}~(0.76)}, \textcolor{darkgreen}{{cn-baike}~(0.27)}, \textcolor{darkgreen}{{mnbvc-news}~(0.10)}, \textcolor{darkgreen}{{cn-book}~(0.24)}, \textcolor{darkgreen}{{cn-legal-case-law}~(0.36)}, \textcolor{darkgreen}{{zhihu-qa}~(0.12)}, \textcolor{navy}{{the-stack-v2}~(4.27)}, \textcolor{navy}{{starcoder}~(2.12)}, \textcolor{navy}{{smollm-python}~(0.20)}, \textcolor{navy}{{mnbvc-code}~(1.13)}, \textcolor{navy}{{yulan-mini-syn-code-inst}~(0.28)}, \textcolor{darkred}{{proof-pile-2}~(1.03)}, \textcolor{darkred}{{automathtext}~(0.68)}, \textcolor{darkred}{{open-web-math-pro}~(0.36)}, \textcolor{darkred}{{cosmopedia}~(0.02)}, \textcolor{darkred}{{fineweb-math}~(0.46)}, \textcolor{darkred}{{dclm-math}~(0.51)}, \textcolor{darkred}{{mathtext}~(0.12)}, \textcolor{darkred}{{basic-math-10m}~(0.04)}, \textcolor{darkred}{{yulan-mini-syn-math-inst}~(0.20)}, \textcolor{darkred}{{yulan-mini-syn-math-doc}~(0.57)}}\\
\addlinespace
\hline
\addlinespace
17 & \newtt{\textcolor{black}{{dclm}~(1.80)}, \textcolor{black}{{fineweb-edu}~(16.20)}, \textcolor{black}{{english-books}~(0.82)}, \textcolor{black}{{pes2o}~(0.80)}, \textcolor{black}{{arxiv}~(1.20)}, \textcolor{black}{{wikipedia}~(0.40)}, \textcolor{black}{{dolma}~(1.24)}, \textcolor{black}{{cosmopedia-v2}~(0.78)}, \textcolor{black}{{cicg-news}~(0.76)}, \textcolor{darkgreen}{{cn-baike}~(0.27)}, \textcolor{darkgreen}{{mnbvc-news}~(0.10)}, \textcolor{darkgreen}{{cn-book}~(0.24)}, \textcolor{darkgreen}{{cn-legal-case-law}~(0.36)}, \textcolor{darkgreen}{{zhihu-qa}~(0.12)}, \textcolor{navy}{{the-stack-v2}~(4.45)}, \textcolor{navy}{{starcoder}~(2.12)}, \textcolor{navy}{{smollm-python}~(0.20)}, \textcolor{navy}{{mnbvc-code}~(1.13)}, \textcolor{navy}{{yulan-mini-syn-code-inst}~(0.28)}, \textcolor{darkred}{{proof-pile-2}~(0.55)}, \textcolor{darkred}{{automathtext}~(0.68)}, \textcolor{darkred}{{cosmopedia}~(0.02)}, \textcolor{darkred}{{fineweb-math}~(0.46)}, \textcolor{darkred}{{dclm-math}~(1.02)}, \textcolor{darkred}{{mathtext}~(0.12)}, \textcolor{darkred}{{basic-math-10m}~(0.04)}, \textcolor{darkred}{{yulan-mini-syn-math-inst}~(0.36)}, \textcolor{darkred}{{yulan-mini-syn-math-doc}~(0.57)}}\\
\addlinespace
\hline
\addlinespace
18 & \newtt{\textcolor{black}{{dclm}~(1.80)}, \textcolor{black}{{fineweb-edu}~(16.20)}, \textcolor{black}{{english-books}~(0.82)}, \textcolor{black}{{pes2o}~(0.80)}, \textcolor{black}{{arxiv}~(1.20)}, \textcolor{black}{{wikipedia}~(0.40)}, \textcolor{black}{{dolma}~(1.24)}, \textcolor{black}{{cosmopedia-v2}~(0.78)}, \textcolor{black}{{cicg-news}~(0.76)}, \textcolor{darkgreen}{{cn-baike}~(0.27)}, \textcolor{darkgreen}{{mnbvc-news}~(0.10)}, \textcolor{darkgreen}{{cn-book}~(0.24)}, \textcolor{darkgreen}{{cn-legal-case-law}~(0.36)}, \textcolor{darkgreen}{{zhihu-qa}~(0.12)}, \textcolor{navy}{{the-stack-v2}~(4.45)}, \textcolor{navy}{{starcoder}~(2.12)}, \textcolor{navy}{{smollm-python}~(0.20)}, \textcolor{navy}{{mnbvc-code}~(1.13)}, \textcolor{navy}{{yulan-mini-syn-code-inst}~(0.31)}, \textcolor{darkred}{{opencoder-llm-math-web}~(1.54)}, \textcolor{darkred}{{automathtext}~(0.00)}, \textcolor{darkred}{{cosmopedia}~(0.02)}, \textcolor{darkred}{{fineweb-math}~(0.17)}, \textcolor{darkred}{{dclm-math}~(0.82)}, \textcolor{darkred}{{mathtext}~(0.12)}, \textcolor{darkred}{{basic-math-10m}~(0.04)}, \textcolor{darkred}{{yulan-mini-syn-math-inst}~(0.52)}, \textcolor{darkred}{{yulan-mini-syn-math-doc}~(0.56)}}\\
\addlinespace
\hline
\addlinespace
19 & \newtt{\textcolor{black}{{dclm}~(1.80)}, \textcolor{black}{{fineweb-edu}~(16.20)}, \textcolor{black}{{english-books}~(0.82)}, \textcolor{black}{{pes2o}~(0.80)}, \textcolor{black}{{arxiv}~(1.20)}, \textcolor{black}{{wikipedia}~(0.40)}, \textcolor{black}{{dolma}~(1.24)}, \textcolor{black}{{cosmopedia-v2}~(0.78)}, \textcolor{black}{{cicg-news}~(0.76)}, \textcolor{darkgreen}{{cn-baike}~(0.02)}, \textcolor{darkgreen}{{mnbvc-news}~(0.02)}, \textcolor{darkgreen}{{cn-book}~(0.24)}, \textcolor{darkgreen}{{cn-legal-case-law}~(0.36)}, \textcolor{darkgreen}{{zhihu-qa}~(0.12)}, \textcolor{navy}{{the-stack-v2}~(4.41)}, \textcolor{navy}{{starcoder}~(2.12)}, \textcolor{navy}{{smollm-python}~(0.20)}, \textcolor{navy}{{mnbvc-code}~(1.06)}, \textcolor{navy}{{opencoder-llm-annealing}~(0.08)}, \textcolor{navy}{{yulan-mini-syn-code-inst}~(0.31)}, \textcolor{darkred}{{opencoder-llm-math-web}~(0.38)}, \textcolor{darkred}{{cosmopedia}~(0.02)}, \textcolor{darkred}{{fineweb-math}~(0.10)}, \textcolor{darkred}{{dclm-math}~(0.82)}, \textcolor{darkred}{{infimm-webmath}~(1.23)}, \textcolor{darkred}{{mathtext}~(0.12)}, \textcolor{darkred}{{basic-math-10m}~(0.04)}, \textcolor{darkred}{{yulan-mini-syn-math-inst}~(0.55)}, \textcolor{darkred}{{yulan-mini-syn-math-doc}~(0.56)}}\\
\addlinespace
\hline
\addlinespace
20 & \newtt{\textcolor{black}{{dclm}~(1.80)}, \textcolor{black}{{fineweb-edu}~(16.20)}, \textcolor{black}{{english-books}~(0.82)}, \textcolor{black}{{pes2o}~(0.80)}, \textcolor{black}{{arxiv}~(1.20)}, \textcolor{black}{{wikipedia}~(0.40)}, \textcolor{black}{{dolma}~(0.84)}, \textcolor{black}{{opencoder-llm-fineweb-corpus}~(0.40)}, \textcolor{black}{{cosmopedia-v2}~(0.78)}, \textcolor{black}{{cicg-news}~(0.76)}, \textcolor{darkgreen}{{mnbvc-news}~(0.02)}, \textcolor{darkgreen}{{cn-book}~(0.26)}, \textcolor{darkgreen}{{cn-legal-case-law}~(0.36)}, \textcolor{darkgreen}{{zhihu-qa}~(0.12)}, \textcolor{navy}{{the-stack-v2}~(4.33)}, \textcolor{navy}{{starcoder}~(2.12)}, \textcolor{navy}{{smollm-python}~(0.20)}, \textcolor{navy}{{mnbvc-code}~(0.92)}, \textcolor{navy}{{opencoder-llm-annealing}~(0.16)}, \textcolor{navy}{{yulan-mini-syn-code-inst}~(0.45)}, \textcolor{darkred}{{opencoder-llm-math-web}~(0.24)}, \textcolor{darkred}{{cosmopedia}~(0.02)}, \textcolor{darkred}{{fineweb-math}~(0.04)}, \textcolor{darkred}{{dclm-math}~(0.56)}, \textcolor{darkred}{{infimm-webmath}~(1.62)}, \textcolor{darkred}{{mathtext}~(0.12)}, \textcolor{darkred}{{basic-math-10m}~(0.01)}, \textcolor{darkred}{{yulan-mini-syn-math-inst}~(0.67)}, \textcolor{darkred}{{yulan-mini-syn-math-doc}~(0.53)}}\\
\addlinespace
\hline
\addlinespace
21 & \newtt{\textcolor{black}{{dclm}~(1.80)}, \textcolor{black}{{fineweb-edu}~(16.20)}, \textcolor{black}{{english-books}~(0.82)}, \textcolor{black}{{pes2o}~(0.80)}, \textcolor{black}{{arxiv}~(1.20)}, \textcolor{black}{{wikipedia}~(0.30)}, \textcolor{black}{{dolma}~(0.84)}, \textcolor{black}{{opencoder-llm-fineweb-corpus}~(0.50)}, \textcolor{black}{{cosmopedia-v2}~(0.78)}, \textcolor{black}{{cicg-news}~(0.76)}, \textcolor{darkgreen}{{mnbvc-news}~(0.02)}, \textcolor{darkgreen}{{cn-book}~(0.26)}, \textcolor{darkgreen}{{cn-legal-case-law}~(0.36)}, \textcolor{darkgreen}{{zhihu-qa}~(0.12)}, \textcolor{navy}{{the-stack-v2}~(4.19)}, \textcolor{navy}{{starcoder}~(2.12)}, \textcolor{navy}{{smollm-python}~(0.20)}, \textcolor{navy}{{mnbvc-code}~(0.80)}, \textcolor{navy}{{opencoder-llm-annealing}~(0.41)}, \textcolor{navy}{{yulan-mini-syn-code-inst}~(0.45)}, \textcolor{darkred}{{opencoder-llm-math-web}~(0.24)}, \textcolor{darkred}{{cosmopedia}~(0.02)}, \textcolor{darkred}{{dclm-math}~(0.23)}, \textcolor{darkred}{{infimm-webmath}~(2.00)}, \textcolor{darkred}{{mathtext}~(0.12)}, \textcolor{darkred}{{yulan-mini-syn-math-inst}~(0.70)}, \textcolor{darkred}{{yulan-mini-syn-math-doc}~(0.53)}}\\
\addlinespace
\hline
\addlinespace
22 & \newtt{\textcolor{black}{{dclm}~(1.80)}, \textcolor{black}{{fineweb-edu}~(16.20)}, \textcolor{black}{{english-books}~(0.82)}, \textcolor{black}{{pes2o}~(0.80)}, \textcolor{black}{{arxiv}~(1.20)}, \textcolor{black}{{dolma}~(0.84)}, \textcolor{black}{{opencoder-llm-fineweb-corpus}~(0.80)}, \textcolor{black}{{cosmopedia-v2}~(0.78)}, \textcolor{black}{{cicg-news}~(0.76)}, \textcolor{darkgreen}{{mnbvc-news}~(0.02)}, \textcolor{darkgreen}{{cn-book}~(0.26)}, \textcolor{darkgreen}{{cn-legal-case-law}~(0.36)}, \textcolor{darkgreen}{{zhihu-qa}~(0.12)}, \textcolor{navy}{{the-stack-v2}~(4.60)}, \textcolor{navy}{{starcoder}~(1.10)}, \textcolor{navy}{{smollm-python}~(0.20)}, \textcolor{navy}{{mnbvc-code}~(1.13)}, \textcolor{navy}{{opencoder-llm-sft-s1}~(0.20)}, \textcolor{navy}{{opencoder-llm-annealing}~(0.38)}, \textcolor{navy}{{yulan-mini-syn-code-inst}~(0.56)}, \textcolor{darkred}{{opencoder-llm-math-web}~(0.24)}, \textcolor{darkred}{{cosmopedia}~(0.02)}, \textcolor{darkred}{{dclm-math}~(0.22)}, \textcolor{darkred}{{infimm-webmath}~(1.98)}, \textcolor{darkred}{{mathtext}~(0.06)}, \textcolor{darkred}{{yulan-mini-syn-math-inst}~(0.78)}, \textcolor{darkred}{{yulan-mini-syn-math-doc}~(0.53)}}\\
\addlinespace
\hline
\addlinespace
23 & \newtt{\textcolor{black}{{dclm}~(1.80)}, \textcolor{black}{{fineweb-edu}~(16.20)}, \textcolor{black}{{english-books}~(0.82)}, \textcolor{black}{{pes2o}~(0.80)}, \textcolor{black}{{arxiv}~(1.20)}, \textcolor{black}{{dolma}~(0.84)}, \textcolor{black}{{opencoder-llm-fineweb-corpus}~(0.80)}, \textcolor{black}{{cosmopedia-v2}~(0.78)}, \textcolor{black}{{cicg-news}~(0.76)}, \textcolor{darkgreen}{{mnbvc-news}~(0.02)}, \textcolor{darkgreen}{{cn-book}~(0.26)}, \textcolor{darkgreen}{{cn-legal-case-law}~(0.36)}, \textcolor{darkgreen}{{zhihu-qa}~(0.12)}, \textcolor{navy}{{the-stack-v2}~(4.16)}, \textcolor{navy}{{starcoder}~(0.80)}, \textcolor{navy}{{mnbvc-code}~(0.85)}, \textcolor{navy}{{opencoder-llm-sft-s1}~(0.20)}, \textcolor{navy}{{opencoder-llm-sft-s2}~(0.15)}, \textcolor{navy}{{opencoder-llm-annealing}~(1.20)}, \textcolor{navy}{{yulan-mini-syn-code-inst}~(0.56)}, \textcolor{darkred}{{opencoder-llm-math-web}~(0.24)}, \textcolor{darkred}{{cosmopedia}~(0.02)}, \textcolor{darkred}{{dclm-math}~(0.22)}, \textcolor{darkred}{{infimm-webmath}~(2.06)}, \textcolor{darkred}{{mathtext}~(0.04)}, \textcolor{darkred}{{lean}~(0.02)}, \textcolor{darkred}{{yulan-mini-syn-math-inst}~(0.92)}, \textcolor{darkred}{{yulan-mini-syn-math-doc}~(0.56)}}\\
\addlinespace
\hline
\addlinespace
24 & \newtt{\textcolor{black}{{dclm}~(1.80)}, \textcolor{black}{{fineweb-edu}~(16.20)}, \textcolor{black}{{english-books}~(0.82)}, \textcolor{black}{{pes2o}~(0.80)}, \textcolor{black}{{arxiv}~(1.20)}, \textcolor{black}{{dolma}~(0.84)}, \textcolor{black}{{opencoder-llm-fineweb-corpus}~(0.80)}, \textcolor{black}{{cosmopedia-v2}~(0.78)}, \textcolor{black}{{cicg-news}~(0.76)}, \textcolor{darkgreen}{{mnbvc-news}~(0.02)}, \textcolor{darkgreen}{{cn-book}~(0.26)}, \textcolor{darkgreen}{{cn-legal-case-law}~(0.36)}, \textcolor{darkgreen}{{zhihu-qa}~(0.12)}, \textcolor{navy}{{the-stack-v2}~(4.07)}, \textcolor{navy}{{starcoder}~(0.80)}, \textcolor{navy}{{mnbvc-code}~(0.85)}, \textcolor{navy}{{opencoder-llm-sft-s1}~(0.25)}, \textcolor{navy}{{opencoder-llm-sft-s2}~(0.08)}, \textcolor{navy}{{opencoder-llm-annealing}~(1.27)}, \textcolor{navy}{{yulan-mini-syn-code-inst}~(0.56)}, \textcolor{darkred}{{opencoder-llm-math-web}~(0.24)}, \textcolor{darkred}{{cosmopedia}~(0.02)}, \textcolor{darkred}{{dclm-math}~(0.19)}, \textcolor{darkred}{{infimm-webmath}~(2.11)}, \textcolor{darkred}{{lean}~(0.04)}, \textcolor{darkred}{{yulan-mini-syn-math-inst}~(0.96)}, \textcolor{darkred}{{yulan-mini-syn-math-doc}~(0.56)}}\\
\addlinespace
\hline
\addlinespace
25 & \newtt{\textcolor{black}{{dclm}~(1.80)}, \textcolor{black}{{fineweb-edu}~(16.20)}, \textcolor{black}{{english-books}~(0.82)}, \textcolor{black}{{pes2o}~(0.80)}, \textcolor{black}{{arxiv}~(1.20)}, \textcolor{black}{{dolma}~(0.84)}, \textcolor{black}{{opencoder-llm-fineweb-corpus}~(0.80)}, \textcolor{black}{{cosmopedia-v2}~(0.78)}, \textcolor{black}{{cicg-news}~(0.76)}, \textcolor{darkgreen}{{mnbvc-news}~(0.02)}, \textcolor{darkgreen}{{cn-book}~(0.26)}, \textcolor{darkgreen}{{cn-legal-case-law}~(0.36)}, \textcolor{darkgreen}{{zhihu-qa}~(0.12)}, \textcolor{navy}{{the-stack-v2}~(4.11)}, \textcolor{navy}{{starcoder}~(0.80)}, \textcolor{navy}{{mnbvc-code}~(0.78)}, \textcolor{navy}{{opencoder-llm-sft-s1}~(0.25)}, \textcolor{navy}{{opencoder-llm-sft-s2}~(0.08)}, \textcolor{navy}{{opencoder-llm-annealing}~(1.30)}, \textcolor{navy}{{ioccc}~(0.00)}, \textcolor{navy}{{yulan-mini-syn-code-inst}~(0.54)}, \textcolor{darkred}{{opencoder-llm-math-web}~(0.24)}, \textcolor{darkred}{{cosmopedia}~(0.02)}, \textcolor{darkred}{{fineweb-math}~(0.20)}, \textcolor{darkred}{{dclm-math}~(0.17)}, \textcolor{darkred}{{infimm-webmath}~(2.11)}, \textcolor{darkred}{{lean}~(0.04)}, \textcolor{darkred}{{yulan-mini-syn-math-inst}~(0.93)}, \textcolor{darkred}{{yulan-mini-syn-math-doc}~(0.45)}}\\
\addlinespace
\hline
\addlinespace
\tabincell{l}{26\\(Decay)} & \newtt{\textcolor{black}{{dclm}~(1.62)}, \textcolor{black}{{fineweb-edu}~(14.58)}, \textcolor{black}{{english-books}~(0.74)}, \textcolor{black}{{pes2o}~(0.72)}, \textcolor{black}{{arxiv}~(1.08)}, \textcolor{black}{{dolma}~(0.48)}, \textcolor{black}{{opencoder-llm-fineweb-corpus}~(1.00)}, \textcolor{black}{{cosmopedia-v2}~(0.70)}, \textcolor{black}{{cicg-news}~(0.68)}, \textcolor{black}{{wizardlm-evol-instruct-v2-196k}~(0.04)}, \textcolor{black}{{less-data}~(0.04)}, \textcolor{black}{{claude-3-opus-claude-3.5-sonnnet-9k}~(0.01)}, \textcolor{black}{{slimorca}~(0.16)}, \textcolor{black}{{tulu-v3.1-mix-preview-4096-olmoe}~(0.20)}, \textcolor{black}{{supernova}~(0.03)}, \textcolor{black}{{magpie-reasoning-150k}~(0.09)}, \textcolor{black}{{spurline}~(0.01)}, \textcolor{black}{{celestia}~(0.04)}, \textcolor{darkgreen}{{mnbvc-news}~(0.01)}, \textcolor{darkgreen}{{cn-book}~(1.00)}, \textcolor{darkgreen}{{zhihu-qa}~(0.05)}, \textcolor{darkgreen}{{chinese-porety}~(0.03)}, \textcolor{navy}{{the-stack-v2}~(3.16)}, \textcolor{navy}{{starcoder}~(0.18)}, \textcolor{navy}{{mnbvc-code}~(0.70)}, \textcolor{navy}{{opencoder-llm-sft-s1}~(0.62)}, \textcolor{navy}{{opencoder-llm-annealing}~(1.09)}, \textcolor{navy}{{magicoder-oss}~(0.06)}, \textcolor{navy}{{textbook-quality-programming}~(0.06)}, \textcolor{navy}{{yulan-mini-syn-code-inst}~(3.00)}, \textcolor{navy}{{code-290k-sharegpt}~(0.08)}, \textcolor{navy}{{evol-codealpaca-v1}~(0.04)}, \textcolor{navy}{{magicoder-evol-instruct-110k}~(0.04)}, \textcolor{navy}{{mathcodeinstruct }~(0.03)}, \textcolor{navy}{{codefeedback-filtered-instruction}~(0.08)}, \textcolor{navy}{{python-code-23k-sharegpt}~(0.01)}, \textcolor{navy}{{evol-instruct-code-80k-v1}~(0.03)}, \textcolor{navy}{{codeexercise-python-27k}~(0.02)}, \textcolor{navy}{{xcoder-80k}~(0.04)}, \textcolor{navy}{{leetcode-solution-python}~(0.00)}, \textcolor{navy}{{tachibana}~(0.03)}, \textcolor{darkred}{{opencoder-llm-math-web}~(0.24)}, \textcolor{darkred}{{cosmopedia}~(0.12)}, \textcolor{darkred}{{infimm-webmath}~(1.33)}, \textcolor{darkred}{{ape210k}~(0.01)}, \textcolor{darkred}{{polytope}~(0.03)}, \textcolor{darkred}{{yulan-mini-syn-math-inst}~(1.44)}, \textcolor{darkred}{{yulan-mini-syn-math-doc}~(0.58)}, \textcolor{darkred}{{mammothmathinstruct}~(0.04)}, \textcolor{darkred}{{openmathinstruct-1}~(0.35)}, \textcolor{darkred}{{fol-nli}~(0.12)}, \textcolor{darkred}{{mathscaleqa-2m}~(0.28)}}\\
\addlinespace
\hline
\addlinespace
\tabincell{l}{27\\(Decay)} & \newtt{\textcolor{black}{{dclm}~(1.44)}, \textcolor{black}{{fineweb-edu}~(12.96)}, \textcolor{black}{{english-books}~(1.48)}, \textcolor{black}{{pes2o}~(0.64)}, \textcolor{black}{{arxiv}~(0.96)}, \textcolor{black}{{dolma}~(0.20)}, \textcolor{black}{{opencoder-llm-fineweb-corpus}~(1.08)}, \textcolor{black}{{cosmopedia-v2}~(0.70)}, \textcolor{black}{{cicg-news}~(0.61)}, \textcolor{black}{{wizardlm-evol-instruct-v2-196k}~(0.04)}, \textcolor{black}{{long-cot}~(0.65)}, \textcolor{black}{{slimorca}~(0.04)}, \textcolor{black}{{tulu-v3.1-mix-preview-4096-olmoe}~(0.25)}, \textcolor{black}{{evolkit-20k}~(0.02)}, \textcolor{black}{{orca-agentinstruct}~(0.49)}, \textcolor{black}{{transcript}~(0.01)}, \textcolor{black}{{spurline}~(0.01)}, \textcolor{black}{{titanium}~(0.02)}, \textcolor{black}{{celestia}~(0.06)}, \textcolor{darkgreen}{{cn-book}~(1.40)}, \textcolor{darkgreen}{{zhihu-qa}~(0.05)}, \textcolor{darkgreen}{{ruozhiba}~(0.00)}, \textcolor{darkgreen}{{chinese-porety}~(0.04)}, \textcolor{navy}{{the-stack-v2}~(1.50)}, \textcolor{navy}{{mnbvc-code}~(0.38)}, \textcolor{navy}{{opencoder-llm-sft-s1}~(0.16)}, \textcolor{navy}{{opencoder-llm-annealing}~(1.13)}, \textcolor{navy}{{magicoder-oss}~(0.11)}, \textcolor{navy}{{textbook-quality-programming}~(0.05)}, \textcolor{navy}{{longwanjuan-github}~(1.82)}, \textcolor{navy}{{yulan-mini-syn-code-inst}~(3.75)}, \textcolor{navy}{{code-290k-sharegpt}~(0.04)}, \textcolor{navy}{{evol-codealpaca-v1}~(0.03)}, \textcolor{navy}{{magicoder-evol-instruct-110k}~(0.03)}, \textcolor{navy}{{mathcodeinstruct }~(0.02)}, \textcolor{navy}{{codefeedback-filtered-instruction}~(0.01)}, \textcolor{navy}{{self-oss-instruct-sc2-exec-filter-50k}~(0.02)}, \textcolor{navy}{{xcoder-80k}~(0.04)}, \textcolor{navy}{{tulu-code}~(0.02)}, \textcolor{navy}{{codefuse-evol-instruct-clean }~(0.03)}, \textcolor{darkred}{{proof-pile-2}~(0.60)}, \textcolor{darkred}{{opencoder-llm-math-web}~(0.45)}, \textcolor{darkred}{{cosmopedia}~(0.02)}, \textcolor{darkred}{{dclm-math}~(0.06)}, \textcolor{darkred}{{infimm-webmath}~(1.34)}, \textcolor{darkred}{{yulan-mini-syn-math-inst}~(1.72)}, \textcolor{darkred}{{yulan-mini-syn-math-doc}~(0.25)}, \textcolor{darkred}{{mammothmathinstruct}~(0.02)}, \textcolor{darkred}{{openmathinstruct-1}~(0.02)}, \textcolor{darkred}{{tulu-math}~(0.14)}, \textcolor{darkred}{{tulu-math-grade}~(0.03)}, \textcolor{darkred}{{tulu-algebra}~(0.02)}, \textcolor{darkred}{{fol-nli}~(0.12)}, \textcolor{darkred}{{reasoning-0.01}~(0.02)}, \textcolor{darkred}{{gretel-math-gsm8k-v1}~(0.01)}, \textcolor{darkred}{{mathscaleqa-2m}~(0.40)}}\\
\addlinespace
\hline
\end{longtable}\label{tab:comp_by_phase}

}






\end{document}